\documentclass{article}

\usepackage{microtype}        

\usepackage{graphicx}         
\usepackage{subcaption} 
\usepackage{booktabs}         
\usepackage{makecell}         
\usepackage{multirow}  
\usepackage[table]{xcolor} 

\usepackage{amsmath}          
\usepackage{amssymb}          
\usepackage{mathtools}        
\usepackage{amsthm}           

\newcounter{mycounter}        
\renewcommand{\themycounter}{\arabic{mycounter}}

\theoremstyle{plain}         

\theoremstyle{definition}

\theoremstyle{remark}

\usepackage[textsize=tiny]{todonotes}  
\usepackage{comment}

\usepackage{xspace}           
\usepackage{enumitem}
\usepackage{array}            
\usepackage{tabularx}
\usepackage{algorithm}
\usepackage{algorithmic}
\usepackage{booktabs}  

\usepackage{hyperref}
\usepackage[capitalize,noabbrev]{cleveref}  

\usepackage[preprint]{icml2026}

\newcommand{\memory}{\emph{External Memory Modules}\xspace}
\newcommand{\memoryone}{\emph{External Memory Module}\xspace}
\newcommand{\system}{\emph{Neuromem}\xspace}

\newcommand{\locomo}{\textsc{LoCoMo}\xspace}
\newcommand{\hustmem}{\textsc{MemoryAgentBench}\xspace}
\newcommand{\membench}{\textsc{MemBench}\xspace}
\newcommand{\longmemeval}{\textsc{LONGMEMEVAL}\xspace}

\newcommand{\tim}{\textsc{TiM}\xspace}
\newcommand{\memorybank}{\textsc{MemoryBank}\xspace}
\newcommand{\memgpt}{\textsc{MemGPT}\xspace}
\newcommand{\amem}{\textsc{A-Mem}\xspace}
\newcommand{\hipporag}{\textsc{HippoRAG}\xspace}
\newcommand{\memoryos}{\textsc{MemoryOS}\xspace}
\newcommand{\ldagent}{\textsc{LD-Agent}\xspace}
\newcommand{\scm}{\textsc{SCM}\xspace}
\newcommand{\memzero}{\textsc{Mem0}\xspace}
\newcommand{\memzerog}{\textsc{Mem0$^{g}$}\xspace}
\newcommand{\secom}{\textsc{SeCom}\xspace}

\renewcommand{\algorithmicrequire}{\textbf{Input:}}
\renewcommand{\algorithmicensure}{\textbf{Output:}}
\newcommand{\algorithmichyper}{\textbf{Hyperparameters:}}
\newcommand{\HYPERPARAMETERS}{\item[\algorithmichyper]}
\renewcommand{\algorithmiccomment}[1]{\hfill \textcolor{gray}{\textit{// #1}}}



\newcommand{\compact}{\vspace{-0pt}}
\newcommand{\subcompact}{\vspace{-0pt}}

\usepackage{marginnote}
\newcommand{\margi}[1]{\marginnote{}}
\newcommand{\margii}[2]{\marginnote{}}

\icmltitlerunning{Neuromem: A Granular Decomposition of the Streaming Lifecycle in External Memory for LLMs}

\begin{document}

\twocolumn[
    \icmltitle{Neuromem: A Granular Decomposition of the Streaming Lifecycle in External Memory for LLMs}
    
    
    
    \icmlsetsymbol{ca}{$\dagger$} 
    
    \begin{icmlauthorlist}
    \icmlauthor{Ruicheng Zhang}{hust}
    \icmlauthor{Xinyi Li}{hust}
    \icmlauthor{Tianyi Xu}{hust}
    \icmlauthor{Shuhao Zhang}{ca,hust}
    \icmlauthor{Xiaofei Liao}{hust}
    \icmlauthor{Hai Jin}{hust}
    
    \end{icmlauthorlist}
    
    \icmlaffiliation{hust}{Cluster and Grid Computing Lab, School of Computer Science and Technology, Huazhong University of Science and Technology, Wuhan, China}
    
    \icmlcorrespondingauthor{Shuhao Zhang}{shuhao\_zhang@hust.edu.cn}
    \icmlkeywords{Large Language Models, Memory Systems, Evaluation Benchmark}
    
    \vskip 0.3in
]



\printAffiliationsAndNotice{\textsuperscript{$\dagger$}Corresponding author.}  

\begin{abstract}

Most evaluations of \memoryone assume a static setting: memory is built offline and queried at a fixed state. In practice, memory is \emph{streaming}: new facts arrive continuously, insertions interleave with retrievals, and the memory state evolves while the model is serving queries. 
In this regime, accuracy and cost are governed by the full memory lifecycle, which encompasses the ingestion, maintenance, retrieval, and integration of information into generation. 
We present \system, a scalable testbed that benchmarks \memory under an interleaved insertion-and-retrieval protocol and decomposes its lifecycle into five dimensions including \emph{memory data structure}, \emph{normalization strategy}, \emph{consolidation policy}, \emph{query formulation strategy}, and \emph{context integration mechanism}. 
Using three representative datasets \locomo, \longmemeval, and \hustmem, \system evaluates interchangeable variants within a shared serving stack, reporting token-level F1 and insertion/retrieval latency.
Overall, we observe that performance typically degrades as memory grows across rounds, and time-related queries remain the most challenging category. The memory data structure largely determines the attainable quality frontier, while aggressive compression and generative integration mechanisms mostly shift cost between insertion and retrieval with limited accuracy gain.

\end{abstract}

\compact

\begin{figure*}[!ht]
    \centering
    \includegraphics[width=\textwidth]{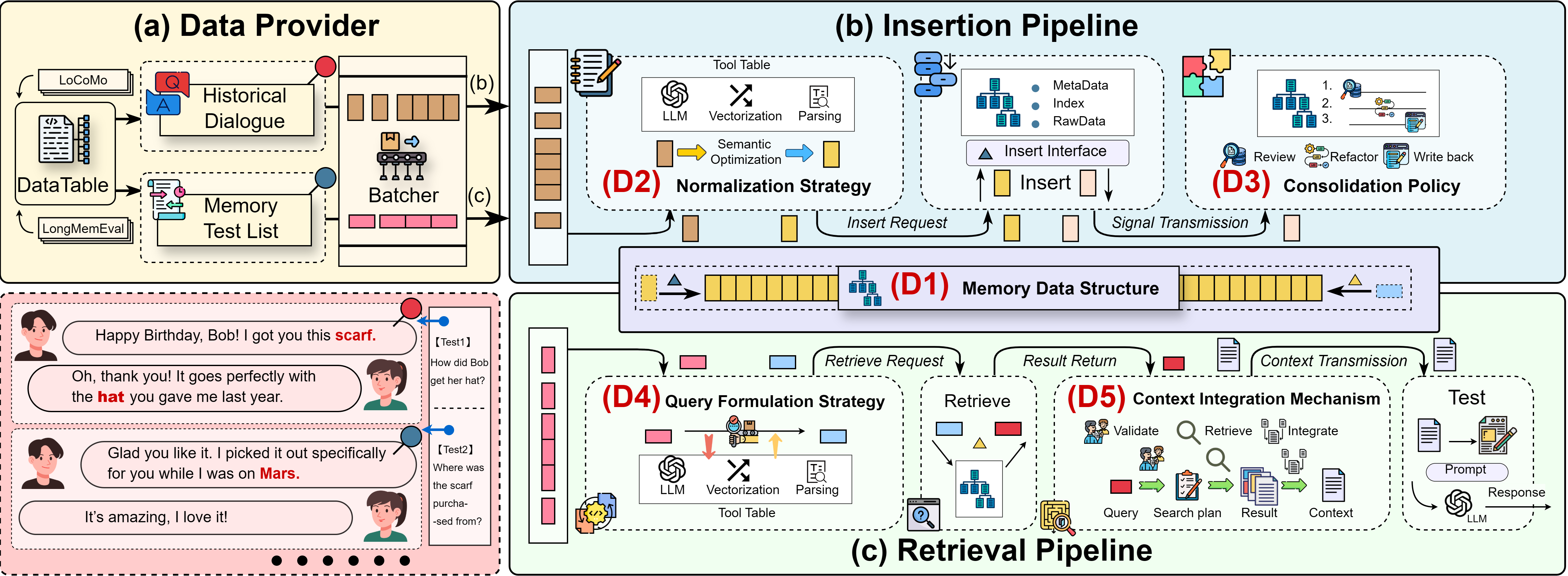}
    \caption{Overview of \system. We decompose the lifecycle into five dimensions (D1--D5) anchored by the (D1) \emph{data structure}. The workflow spans the (b) \textbf{Insertion Pipeline} (D2 \emph{normalization}, D3 \emph{consolidation}) and (c) \textbf{Retrieval Pipeline} (D4 \emph{query formulation}, D5 \emph{context integration}), enabling controlled ablations under an interleaved insertion and retrieval protocol.}
    \label{fig:overview}
    \vspace{-15pt}
\end{figure*}

\section{Introduction}
\label{sec:intro}

\memory are becoming a core building block for long-horizon, interactive LLM systems, including medical assistants~\cite{YUAN2024100030}, embodied agents~\cite{liang2025largemodelempoweredembodied, salama2025meminsightautonomousmemoryaugmentation, long2025seeinglisteningrememberingreasoning}, and educational tutoring~\cite{pan2025memoryconstructionretrievalpersonalized, emotionalrag, salemi2024optimizationmethodspersonalizinglarge}. By persisting, updating, and selectively retrieving information beyond the context window, \memory provide access to relevant prior facts during reasoning and enable cross-session continuity. 
Unlike \emph{parametric} memory~\cite{behrouz2024titanslearningmemorizetest, yan2026memoryr1enhancinglargelanguage, yu2026agenticmemorylearningunified}, where static knowledge is integrated into model weights via fine-tuning or continual learning, \memory maintain state in an explicit, editable store decoupled from the backbone model.
While this separation enables continual updates without model modification, it transforms memory into a complex system component requiring insertion and retrieval pipelines designed under strict latency constraints.

Today, driven by diverse application demands and deployment budgets, \memory have expanded into a large and fast-evolving design space~\cite{madaan-etal-2022-memory, dalvi-mishra-etal-2022-towards, emotionalrag, gutierrez2025hipporag, ong-etal-2025-towards, xu2025amemagenticmemoryllm, latimer2025hindsight2020buildingagent, li2025helloagainllmpoweredpersonalized, huang2026licomemorylightweightcognitiveagentic, chen2026telemembuildinglongtermmultimodal, tao2026memboxweavingtopiccontinuity}. Yet, choosing an appropriate design for a concrete workload remains difficult. 
Furthermore, memory is \emph{streaming} in practice: new facts arrive continuously, insertions interleave with retrievals, and the memory state evolves while the model is serving queries~\cite{streaming}. 
This interleaving exposes non-obvious trade-offs and reveals how computational costs are shifted between insertion and retrieval~\cite{huaweimemory}, but existing evaluations~\cite{locomo,longbench, membench} rarely make it clear which design decisions along the memory lifecycle drive the observed accuracy--latency outcomes.

Recent benchmark efforts have begun to move beyond single-number leaderboards. LoCoMo and MemBench~\cite{locomo, membench} emphasize long-horizon interactions and introduce efficiency measurements, while Minerva~\cite{xia2025minerva} takes a complementary angle: it programmatically generates fine-grained, interpretable \emph{in-context} memory tests to diagnose what language models can do with their prompt memory. These are valuable foundations, but critical gaps remain for understanding \memory in realistic deployments. First, many comparisons remain coarse-grained and end-to-end, ranking whole systems and obscuring the impact of specific design decisions. Second, efficiency reporting is often under-specified: average read/write times conceal where overhead is incurred, and system-level metrics (e.g., memory footprint) and attribution between memory-module overhead and model inference are frequently missing. Third, many evaluations still operate in a static or quasi-static regime (offline build, then query), whereas real deployments are dominated by interleaved insertion and retrieval, where interference accumulates as memory grows and maintenance competes with serving latency.

To capture these dynamics, we argue that evaluation must be both \emph{streaming} and \emph{lifecycle-aware}. Streaming evaluation reflects the interleaved insertion and retrieval regime in which memory evolves over time. Complementarily, lifecycle-aware evaluation addresses the attribution challenge by dissecting the monolithic memory system into granular stages. To operationalize this, we decompose \memory into five dimensions covering the full data lifecycle: (D1) \emph{memory data structure}, (D2) \emph{normalization strategy}, (D3) \emph{consolidation policy}, (D4) \emph{query formulation strategy}, and (D5) \emph{context integration mechanism} (Figure~\ref{fig:overview}).

We instantiate this framework in \system, a scalable testbed that benchmarks \memory under the interleaved insertion and retrieval protocol. By isolating design dimensions within a shared serving stack, \system enables precise attribution of performance gains.
Using three representative datasets \locomo, \longmemeval and \hustmem, \system evaluates interchangeable variants, reporting token-level F1 together with insertion/retrieval latency.
Our analysis reveals a conservation of complexity: systems typically shift computational debt between insertion and retrieval rather than eliminating it. We find that hybrid data structures (D1) set the strict accuracy ceiling, while expensive operations like summarization and multi-query fusion often act as latency traps yielding negligible gains. Finally, performance degrades as memory accumulates, with temporal reasoning remaining a persistent bottleneck.


\textbf{Contributions.} This paper makes three contributions. (1) We introduce \system, an open-source testbed for evaluating \memory under a streaming protocol within a unified serving stack. (2) We propose a granular decomposition of memory architectures into five design dimensions and implement representative design variants for each, enabling precise, controlled ablations. (3) We conduct an extensive empirical study across diverse long-horizon benchmarks to distill practical design guidelines on balancing reasoning quality with insertion and retrieval costs.


\textbf{Paper organization.} Section~\ref{sec:related_work} reviews related work on external memory modules and evaluation benchmarks. Section~\ref{sec:design_of_EMM} formalizes the problem setting and notation, and decomposes the memory lifecycle into five design dimensions (D1--D5). Section~\ref{sec:methodology} introduces the interleaved streaming evaluation protocol, details the system instantiations mapped to our taxonomy, and describes the metrics and experimental platform. Section~\ref{sec:evaluation} presents the experimental analysis and key findings, and Section~\ref{sec:conclusion} concludes with implications and future directions.
\compact

\vspace{-0.3cm}
\section{Related Work}
\label{sec:related_work}

\subsection{External Memory Modules}
To address the fundamental limitations of \textbf{statelessness} and \textbf{finite context windows}, which confine models to ephemeral interactions and thereby preclude long-horizon coherence, personalized adaptation, and continuous self-evolution, a wide array of \memory has been proposed.  These systems serve as a bridge for long-horizon continuity, ranging from general-purpose retrieval tools~\cite{ kang2025memoryosaiagent, zhong2023memorybankenhancinglargelanguage} to specialized agentic frameworks~\cite{zhang-etal-2024-llm-based, appmemory}; a comprehensive review is provided in Appendix~\ref{app:external_memory_modules}. Despite this diversity, current designs remain predominantly \textit{monolithic}: they tightly couple storage with maintenance logic into ``black box" systems, obscuring the contribution of individual lifecycle stages and hindering optimization for real-time interaction.

\subsection{Evaluation Benchmarks}

Parallel to the rapid evolution of \memory, the evaluation landscape has made significant strides in expanding task diversity and reasoning complexity, as exemplified by \locomo, \longmemeval, and \hustmem (see Appendix~\ref{app:memory_benchmark}). Complementing these accuracy-focused benchmarks, pioneering efforts like \textsc{MemBench} have also made valuable contributions by incorporating efficiency metrics into the evaluation loop. However, even these advancements are limited by their reliance on isolated latency measurements and static, retrospective protocols; this inability to capture the interleaved dynamics of continuous state evolution motivates our proposal for a lifecycle-aware streaming evaluation.








\compact

\section{Design of External Memory Module}
\label{sec:design_of_EMM}

\subsection{Problem Setting}
We formalize the \memoryone as a stateful component processing a continuous stream of operations. 
Let $\mathcal{R} = \{ r_i = (\tau_i, \textsc{type}_i, \textsc{payload}_i) \}_{i=1}^\infty$ denote the request stream, where each request comprises a timestamp $\tau_i$, an operation type $\textsc{type}_i \in \{\textsc{Insert}, \textsc{Retrieve}\}$, and a payload.
The payload is denoted as context $h$ for \textsc{Insert} operations and query $q$ for \textsc{Retrieve} operations.

The memory maintains a state sequence $\{M^{(k)}\}_{k=0}^\infty$ and manages its lifecycle via two primary pipelines:

\paragraph{Insertion Pipeline.}
The state $M^{(k)}$ is derived from the previous state by processing the $k$-th \textsc{Insert} request with context $h^{(k)}$:
\begin{equation}
M^{(k)} = \textsc{PostIns}\bigl( M^{(k-1)},\; \textsc{PreIns}(h^{(k)}) \bigr), \label{eq:insert}
\end{equation}
where $\textsc{PreIns}$ normalizes the raw context for insertion, and $\textsc{PostIns}$ updates the memory state according to maintenance policies.

\paragraph{Retrieval Pipeline.}
For a \textsc{Retrieve} request with query $q$, the system accesses the current memory state $M^{(k^*)}$ (where $k^*$ denotes the state after the most recent insertion):
\begin{equation}
c = \textsc{PostRet}\bigl( M^{(k^*)},\; \textsc{PreRet}(q) \bigr), \label{eq:retrieve}
\end{equation}
where $\textsc{PreRet}$ formulates the retrieval signal from the user query, and $\textsc{PostRet}$ synthesizes the final context by retrieving and refining evidence to construct the final context $c$.

These four operators, anchored by the underlying structure of $M$, constitute the functional backbone of any \memoryone. To isolate the impact of specific architectural decisions, we map these mathematical components to five orthogonal design dimensions (D1--D5) in the following taxonomy.

\subsection{Design Aspects of External Memory Module}
\label{subsec:design_aspects}

Implementation decisions for the operators in Eq.~(\ref{eq:insert})--(\ref{eq:retrieve}) jointly determine system performance. We identify five orthogonal dimensions (D1--D5) to decompose this design space(see Appendix~\ref{app:design_aspects_of_EEMs} for more details):

\textbf{(D1) Memory Data Structure ($M$):} This dimension functions as the primary storage substrate, organizing incoming information into queryable units.The fundamental design choice lies in \textit{Topology}: \textit{Partitional} architectures store memories as discrete, independent chunks (e.g., flat vector stores), prioritizing fast similarity search; whereas \textit{Hierarchical} architectures model explicit dependencies between entities (e.g., knowledge graphs), enabling multi-hop reasoning. Within the \memoryone lifecycle,  this structure governs the overhead of state updates and the physical efficiency of subsequent retrievals.

\textbf{(D2) Normalization ($\textsc{PreIns}$) \& (D3) Consolidation ($\textsc{PostIns}$):}
In conjunction with D1, these dimensions realize the \textit{Insertion Pipeline}, defining how memory state evolves over time. 
Specifically, \textbf{D2} performs preprocessing to translate unstructured history into the discrete storable units. This stage determines the granularity and fidelity of the ingested information.
Following insertion, \textbf{D3} manages the \textit{state evolution} of $M$ to ensure long-horizon stability and compactness. 
It implements maintenance policies such as Conflict Resolution, Decay Eviction for capacity control, and Structural Refinement for hierarchical memories. 
Together, D2 and D3 define the computational cost and representational quality of the memory's insertion pipeline.

\textbf{(D4) Query Formulation ($\textsc{PreRet}$) \& (D5) Context Integration ($\textsc{PostRet}$):} 
Operating upon the storage substrate D1, these dimensions operationalize the \textit{Retrieval Pipeline}, governing how stored context is surfaced. 
Functionally, \textbf{D4} acts as the alignment interface, bridging user intent with retrieval signals compatible with D1, such as generating embeddings for partitional stores or decomposing paths for hierarchical graphs. 
Upon retrieving evidence, \textbf{D5} serves as the \textit{output synthesizer} that executes the $\textsc{PostRet}$ operator, refining raw candidates into the final context via Reranking or Fusion.
Ultimately, the synergy between D4 and D5 dictates the retrieval precision and the contextual coherence of the generated response.

By isolating these dimensions, \system allows us to measure how computational cost is shifted between insertion and retrieval pipelines.

\compact

\section{Methodology}
\label{sec:methodology}
We present our streaming evaluation framework, detailing the protocol and workload adaptation (\S\ref{sec:streaming_protocol}), representative systems mapped to the taxonomy (\S\ref{sec:system_instantiations}), and the metrics and testbed (\S\ref{sec:metrics_platform}).

\subsection{Streaming Protocol and Workloads}
\label{sec:streaming_protocol}

To model realistic deployment, the request stream $\mathcal{R}$ is instantiated as a strictly ordered time-series sequence. Unlike static benchmarks that decouple memory construction from usage, we enforce causality by sorting requests $r_i$ based on timestamps $\tau_i$. The system processes this stream sequentially: for an \textsc{Insert} request, it invokes Eq.~\eqref{eq:insert} to evolve the state from $M^{(k-1)}$ to $M^{(k)}$; for a \textsc{Retrieve} request, it executes Eq.~\eqref{eq:retrieve} against the \emph{current} state $M^{(k)}$. This interleaved execution ensures that a query $q_t$ is strictly limited to accessing evidence from contexts $\{h_{t'} \mid \tau_{t'} < \tau_t\}$ already integrated into the memory. By adhering to this temporal order, the protocol naturally prevents future leakage without requiring artificial visibility masks, while explicitly capturing the computational costs of state transitions.

\locomo, \longmemeval, and \hustmem are serialized into the stream $\mathcal{R}$ to preserve chronological ordering. Evaluation is triggered at fixed intervals (e.g., 20\%) to capture performance evolution as memory accumulates, rather than only at the final state. Preprocessing details are in Appendix~\ref{app:dataset_details}.

\subsection{System Instantiations}
\label{sec:system_instantiations}
We reproduce representative memory modules by explicitly mapping their components to our D1--D5 taxonomy, covering \tim~\cite{liu2023thinkinmemoryrecallingpostthinkingenable}, \memorybank~\cite{zhong2023memorybankenhancinglargelanguage}, \memgpt~\cite{packer2024memgptllmsoperatingsystems}, \amem~\cite{xu2025amemagenticmemoryllm}, \hipporag (and HippoRAG 2)~\cite{gutierrez2025hipporag}, \memoryos~\cite{kang2025memoryosaiagent}, \ldagent~\cite{li2025helloagainllmpoweredpersonalized}, \scm~\cite{wang2024enhancing}, \memzero and \memzerog~\cite{mem0}, and \secom~\cite{pan2025memoryconstructionretrievalpersonalized}. This decomposition transforms monolithic systems into atomic operators, allowing us to construct interchangeable variants and conduct granular ablations to isolate the impact of specific design decisions. Implementation details are provided in Appendix~\ref{app:design_aspects_of_EEMs}.

\subsection{Metrics and Platform}
\label{sec:metrics_platform}

We assess reasoning accuracy via Token-level F1 with Porter stemming, while quantifying system efficiency through granular Insertion Latency and Retrieval Latency. All experiments are executed on a unified serving stack hosting \texttt{pangu-1b} on Huawei Ascend 910B NPUs and \texttt{Llama-3.1-8B} on Nvidia A6000 GPUs, using asynchronous scoring to avoid latency interference.
\compact

\section{Experimental Analysis}
\label{sec:evaluation}
This section evaluates \memory under the \emph{streaming} protocol (Section~\ref{sec:streaming_protocol}) using \system. The goal is attribution: we quantify how each lifecycle dimensions (D1--D5) influences answer quality and system cost as memory evolves, rather than treating implementations as black boxes.

\memory are highly sensitive to both operational hyperparameters (e.g., retrieval top-$k$) and workload characteristics~\cite{packer2024memgptllmsoperatingsystems}. 
To enable deep attribution without combinatorial explosion, \system focuses the full D1--D5 ablation on \locomo, while using \longmemeval and \hustmem to cross-validate structural trends (D1). 
We first distill our primary empirical observations in Section~\ref{subsec:findings}, followed by a granular analysis of each design dimension (D1--D5) via controlled ablations in Sections~\ref{subsec:memory_data_structure}--\ref{subsec:context_integration_mechanism}. 
Finally, we delineate the computational and theoretical boundaries of the current study in Section~\ref{subsec:limitation}.

\subcompact
\begin{figure*}[t]
    \centering
    \includegraphics[height=6cm,width=\linewidth]{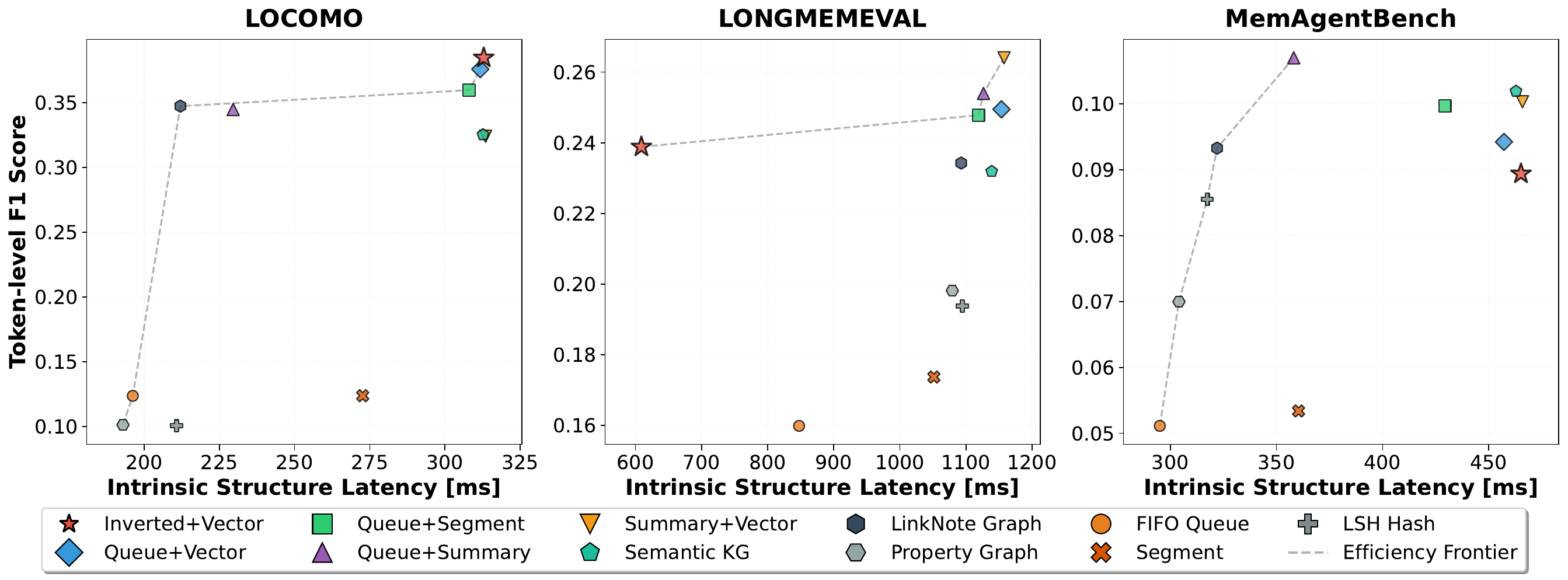}
    \caption{\textbf{Intrinsic Efficiency Frontier.} Plotting accuracy against \textbf{Intrinsic Structure Latency}. The \textbf{Inverted+Vector} anchors the Pareto frontier on reasoning benchmarks (Left, Center), whereas the trade-off inverts on maintenance-heavy tasks (Right) where lightweight queues achieve superior efficiency.}
    \label{fig:unified_frontier}
    \vspace{-15pt}
\end{figure*}

\subsection{Findings}
\label{subsec:findings}

We distill the longitudinal evaluation results into five core empirical findings that govern the accuracy-latency trade-offs in streaming memory systems. Detailed decompositions supporting these conclusions are provided in subsequent sections.

\noindent\textbf{[F1] Temporal Entropy is Inherent.}
Performance universally degrades as interaction history accumulates, establishing a baseline of noise that persists across architectures.
Crucially, computationally expensive consolidation policies driven by LLMs fail to mitigate this decay, exhibiting degradation rates comparable to unmaintained baselines.
These proactive methods exhibit the exact same degradation rate of approximately 22\% as unmaintained baselines.
This finding confirms that resolving semantic contradictions at ingestion time acts as a premature optimization that incurs significant latency costs without countering the entropy inherent to long-horizon interaction.

\noindent\textbf{[F2] Storage Architecture Dictates Performance Bounds.}
The underlying memory data structure establishes the hard ceiling on reasoning quality.
Multi-layer architectures that combine lexical and semantic indexing, typified by the \texttt{Inverted+Vector} design, consistently dominate single-signal baselines.
These hybrid systems justify their intrinsic latency by anchoring the efficiency frontier, whereas simpler approaches fail to capture necessary context.
Our ablations prove that downstream optimizations in query formulation or context integration cannot compensate for information loss incurred at the storage level.

\noindent\textbf{[F3] Semantic Compression is Lossy.}
Aggressive abstraction during the normalization phase is consistently destructive.
Transforming natural dialogue into rigid structured schemas precipitates functional collapse, reducing F1 scores by over 50\%. Our results demonstrate that structural schemas discard subtle linguistic context and temporal markers essential for effective retrieval, rendering the preservation of raw textual texture the strictly superior strategy.

\noindent\textbf{[F4] Generative Optimization incurs a Latency Tax.}
Incorporating generative steps into the retrieval pipeline imposes a prohibitive latency penalty with diminishing returns.
Strategies such as query decomposition and multi-query expansion inflate processing times by an order of magnitude, often exceeding one second per turn.
Despite this high computational cost, these methods yield negligible or even negative accuracy gains compared to direct processing.
This inverse cost-benefit profile persists even when scaling to stronger backbones like Llama-3-8B, indicating a fundamental inefficiency in generative retrieval enhancement.

\noindent\textbf{[F5] Heuristics Define the Efficiency Frontier.}
In the constraint-heavy regime of online streaming, deterministic heuristics consistently outperform generative interventions.
Mechanisms like heat-based migration and heuristic context augmentation achieve parity or superiority in reasoning accuracy compared to their model-based counterparts.
Crucially, these heuristic methods operate with negligible latency overheads often measuring less than one millisecond.
This establishes that the optimal design strategy for streaming memory is to effectively decouple state maintenance from expensive token generation processes.

\subsection{Memory Data Structure}
\label{subsec:memory_data_structure}

We analyze the impact of memory representations (D1) by instantiating the core data structures from the representative systems described in \S\ref{sec:system_instantiations}.
These implementations span the Partitional and Hierarchical paradigms, extending from the Fifo Queue typified by \scm to the Linknote Graph underpinning \amem.
To evaluate the intrinsic properties of these storage substrates, we integrate them as interchangeable modules within a unified serving pipeline, fixing all other dimensions (D2--D5) to minimal baselines.
Figure~\ref{fig:unified_frontier} visualizes the resulting accuracy--latency landscape across the three workloads.

\begin{table}[h]
    \centering
    \caption{\textbf{Temporal Degradation on LoCoMo.} Performance consistently declines across all structures as interaction history accumulates (R1$\rightarrow$R5).}
    \label{tab:locomo_evolution}
    \resizebox{\columnwidth}{!}{
    \begin{tabular}{l|ccccc|c}
        \toprule
        \multirow{2}{*}{\textbf{Structure Type}} & \multicolumn{5}{c|}{\textbf{Token-level F1 per Round}} & \textbf{Degradation} \\
        & \textbf{R1} & \textbf{R2} & \textbf{R3} & \textbf{R4} & \textbf{R5} & \textbf{($\Delta$ R1$\rightarrow$R5)} \\
        \midrule
        Fifo Queue & 0.169 & 0.128 & 0.118 & 0.109 & 0.094 & \textbf{-44.4\%} \\
        Queue+Segment & 0.395 & 0.362 & 0.356 & 0.349 & 0.338 & -14.4\% \\
        \textbf{Inverted+Vector} & \textbf{0.411} & \textbf{0.395} & \textbf{0.375} & \textbf{0.385} & \textbf{0.358} & \textbf{-12.9\%} \\
        \bottomrule
    \end{tabular}
    }
    \vspace{-10pt}
\end{table}

\noindent\textbf{O\themycounter\stepcounter{mycounter}: Performance degradation due to round accumulation is a universal law of \memoryone.}
Our longitudinal analysis in Table~\ref{tab:locomo_evolution} reveals consistent performance degradation as interaction history accumulates.
From Round 1 to Round 5, even the robust \texttt{Inverted+Vector} structure suffers an approximate 13\% decline as the F1 score falls from 0.411 to 0.358.
In comparison, limited-capacity baselines like the \texttt{Fifo queue} deteriorate by over 44\% over the same period.
This inverse correlation between history length and retrieval precision underscores that structural optimization alone cannot arrest the noise accumulation inherent to streaming, necessitating active consolidation policies.

\noindent\textbf{O\themycounter\stepcounter{mycounter}: Multi-layer Partitional structures optimize the Intrinsic Return on Investment (ROI).}
\noindent Visualizing the accuracy--latency landscape (Figure~\ref{fig:unified_frontier}) exposes distinct cost--benefit profiles.
For reasoning benchmarks such as \textsc{LoCoMo} and \textsc{LongMemEval}, the Multi-layer \texttt{Inverted+Vector} design justifies its higher intrinsic latency.
It anchors the efficiency frontier by converting this computational investment into essential resilience to surpass Hierarchical graphs in recall while avoiding the prohibitive overhead associated with generative baselines.
However, this return on investment inverts in high-churn environments like \textsc{MemAgentBench}.
In these volatile streams, the maintenance overhead of vector indices becomes a liability, rendering the Single-layer \texttt{Queue+Summary} superior in both accuracy and speed.
Consequently, Multi-layer designs emerge as the optimal choice for reasoning stability, whereas Single-layer approaches dominate in scenarios requiring frequent maintenance.

\begin{figure*}[t]
    \centering
    {
    \begin{subfigure}[t]{0.245\linewidth}
        \raisebox{5pt}{
        \includegraphics[width=\linewidth]{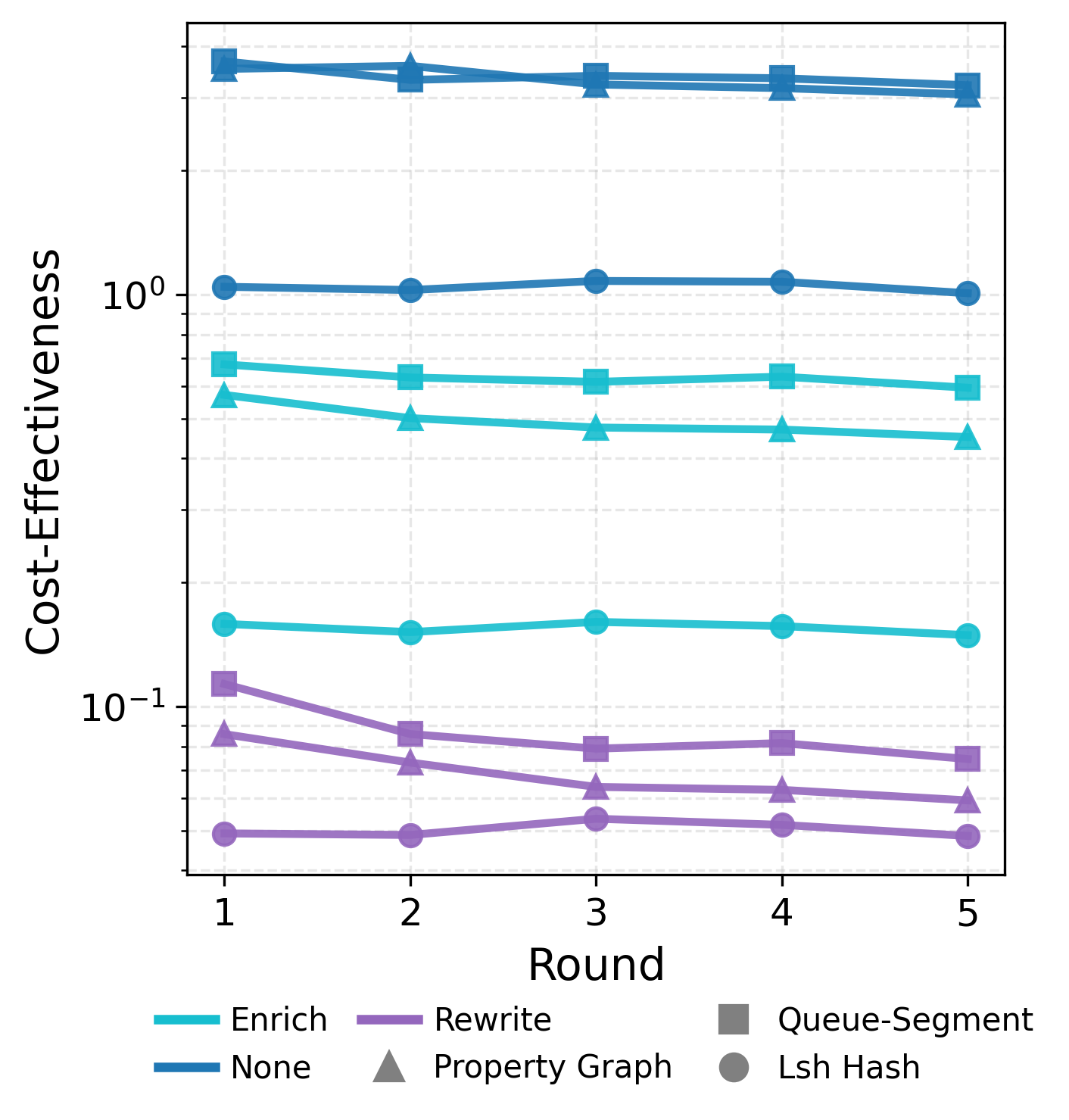}}
        \caption{Normalization Strategy}
    \end{subfigure}%
    }
    \hfill
    \begin{subfigure}[t]{0.255\linewidth}
        \includegraphics[width=\linewidth]{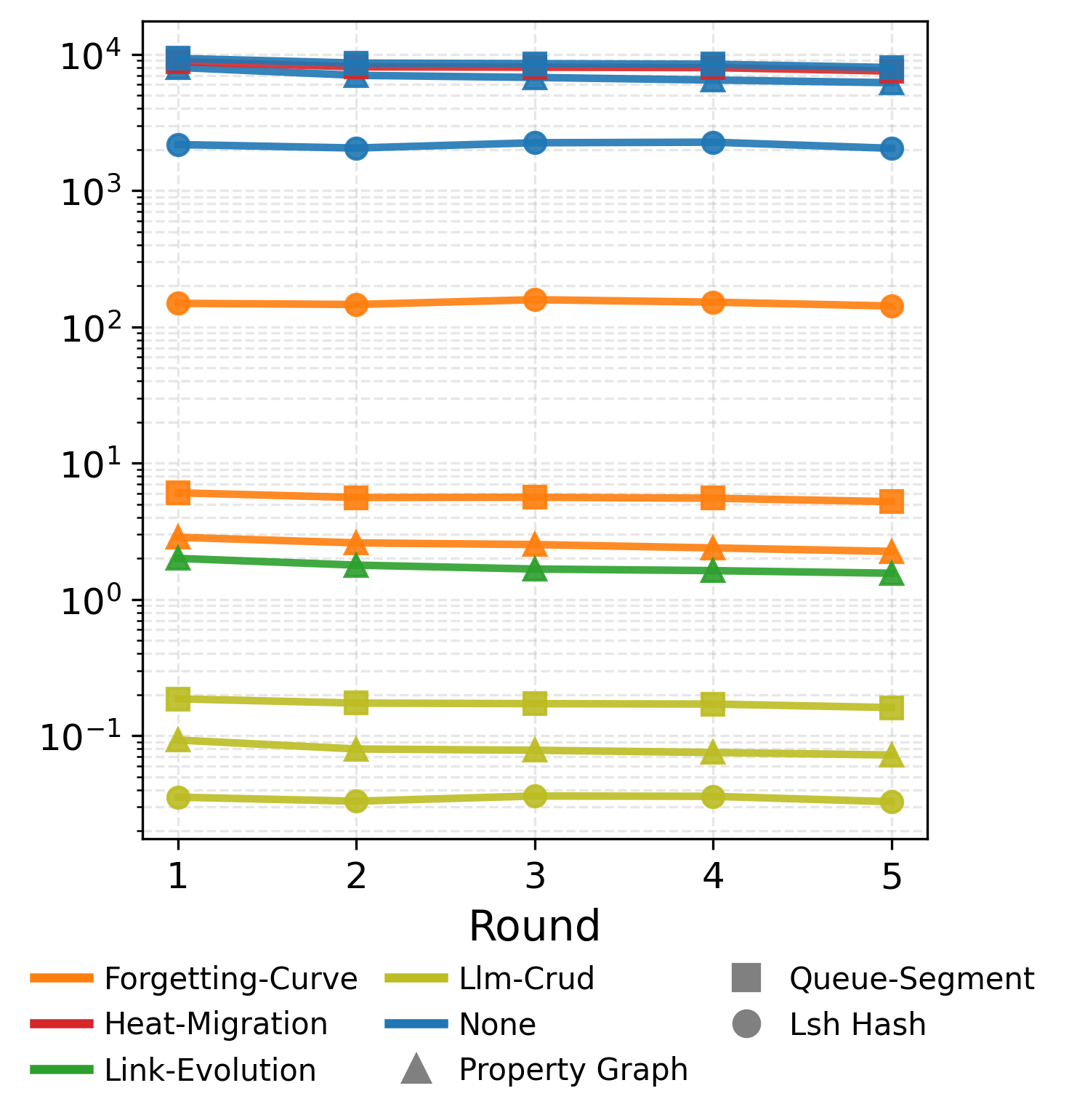}
        \caption{Consolidation Policy}
    \end{subfigure}%
    \hfill
    \begin{subfigure}[t]{0.245\linewidth}
        \includegraphics[width=\linewidth]{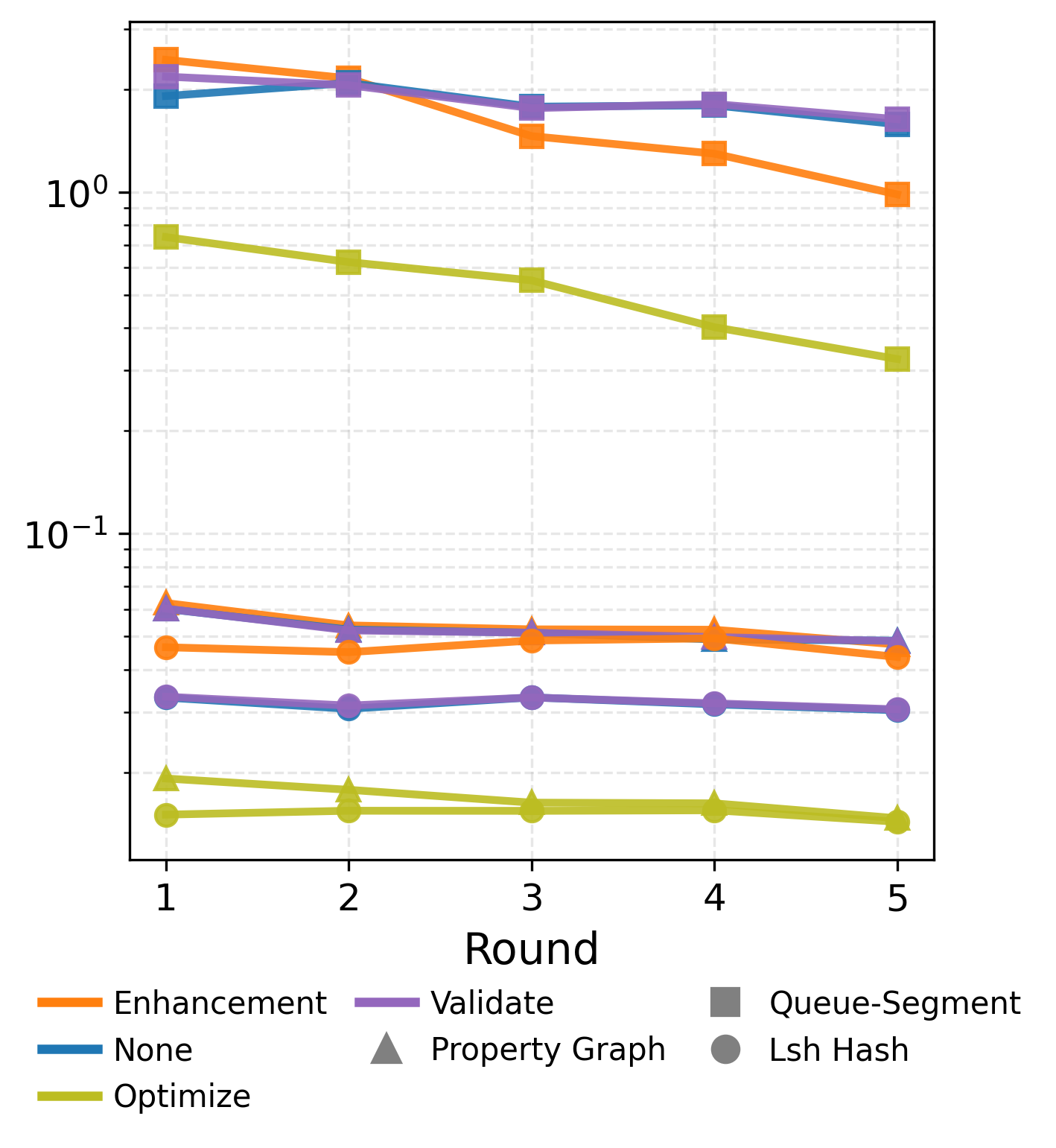}
        \caption{Query Formulation}
    \end{subfigure}%
    \hfill
    \begin{subfigure}[t]{0.235\linewidth}
        \includegraphics[width=\linewidth]{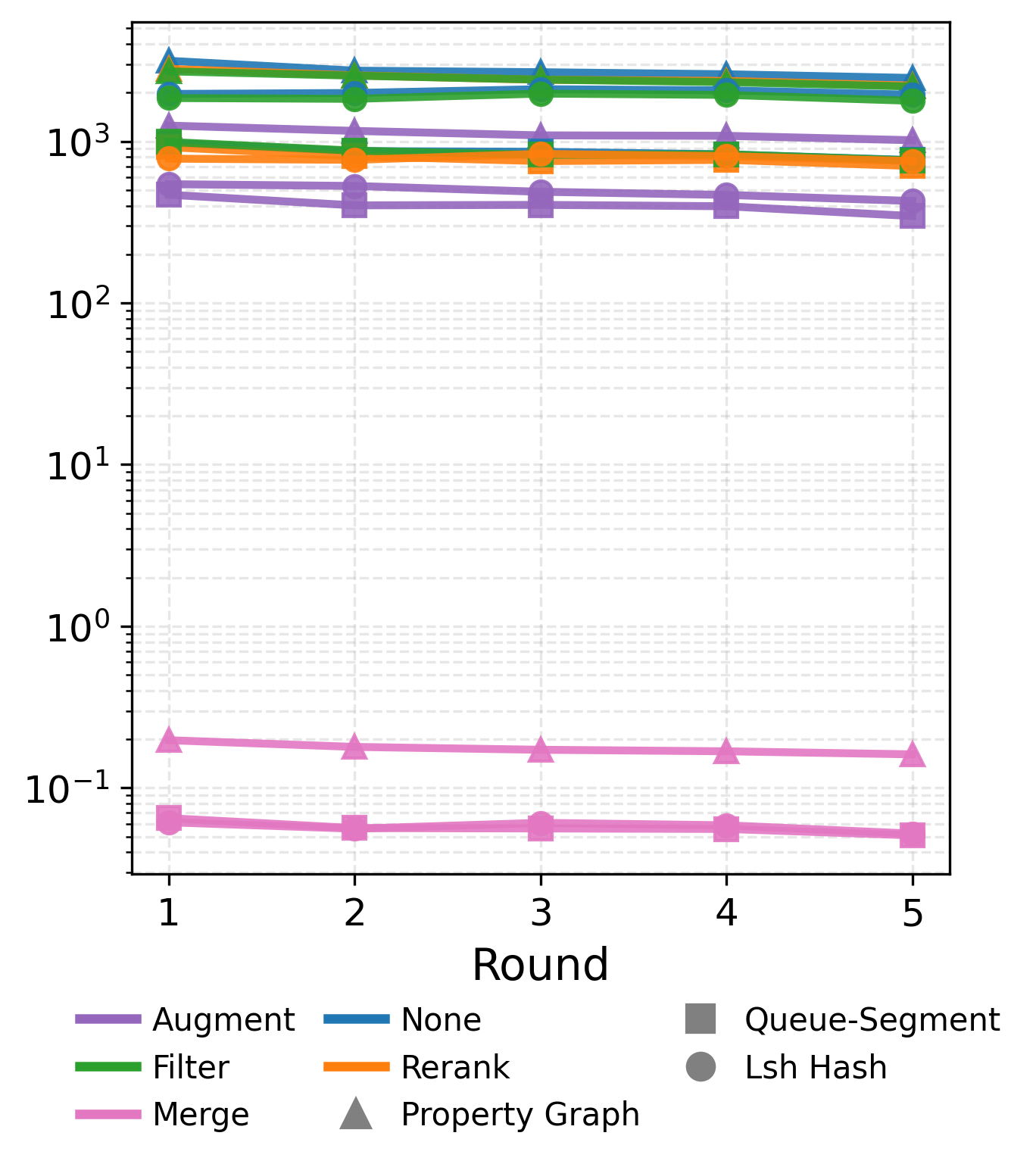}
        \caption{Context Integration}
    \end{subfigure}%
    \caption{\textbf{Breakdown of Cost-Effectiveness Evolution by Memory Lifecycle Dimension.} The plots illustrate the F1-per-second trajectory across five rounds for varying strategies in D2--D5.}
    \label{fig:four_subfigures}
    \vspace{-15pt}
\end{figure*}

\subsection{Normalization Strategy}
\label{subsec:normalization_strategy}

We anchor our evaluation on three distinct storage architectures:LSH Hash, Queue+Segment and Property Graph. These configurations, which instantiate the representative design paradigms of \memoryos, \memzerog, and \tim respectively, serve as the default baselines for this and subsequent dimensions unless otherwise specified.

We investigate three normalization strategies, selected to represent the most common preprocessing paradigms: \texttt{None} preserves the raw context; \texttt{Enrich} performs summarization; and \texttt{Rewrite} extracts triplets.
Figure~\ref{fig:four_subfigures}.(a) illustrates the resulting Cost-Effectiveness evolution across rounds.

\begin{table}[h]
    \centering
    \caption{\textbf{Impact of Normalization on LoCoMo (Round1-5).} While \texttt{Rewrite} causes accuracy collapse, \texttt{Enrich} incurs massive insertion costs for transient or negligible gains. Best F1 scores are bolded.}
    \label{tab:normalization_impact}
    \resizebox{\linewidth}{!}{
    \begin{tabular}{llcccccc}
        \toprule
        \multirow{2}{*}{\textbf{Backend}} & \multirow{2}{*}{\textbf{Strategy}} & \multicolumn{5}{c}{\textbf{Token-level F1 per Round}} & \multirow{2}{*}{\textbf{Latency}} \\
        \cmidrule(lr){3-7}
         & & \textbf{R1} & \textbf{R2} & \textbf{R3} & \textbf{R4} & \textbf{R5} \\
        \midrule
        \multirow{3}{*}{Queue+Segment} 
        & \texttt{None} & \textbf{0.371} & \textbf{0.335} & \textbf{0.342} & \textbf{0.338} & \textbf{0.325} & \textbf{146\,ms} \\
        & \texttt{Enrich} & 0.369 & 0.343 & 0.334 & 0.344 & 0.324 & 580\,ms \\
        & \texttt{Rewrite} & 0.171 & 0.129 & 0.119 & 0.123 & 0.112 & 1552\,ms \\
        \midrule
        \multirow{3}{*}{Property Graph} 
        & \texttt{None} & 0.347 & 0.353 & 0.318 & 0.311 & \textbf{0.301} & \textbf{1171\,ms} \\
        & \texttt{Enrich} & \textbf{0.380} & 0.334 & 0.317 & 0.313 & 0.300 & 2567\,ms \\
        & \texttt{Rewrite} & 0.163 & 0.139 & 0.121 & 0.119 & 0.113 & 4491\,ms \\
        \bottomrule
    \end{tabular}
    }
    \vspace{-10pt}
\end{table}

\noindent\textbf{O\themycounter\stepcounter{mycounter}: Aggressive restructuring is reliably destructive.}
Transforming dialogue into rigid triples using the \texttt{Rewrite} strategy causes immediate functional collapse. 
As detailed in Table~\ref{tab:normalization_impact}, the F1 score for Queue+Segment drops from 0.371 with the \texttt{None} baseline to 0.171 using \texttt{Rewrite} in the first round, and this deficit widens by Round 5 where accuracy falls from 0.325 down to 0.112. 
Temporal reasoning suffers the most with performance plummeting to a near-random F1 level of approximately 0.033. 
This confirms that semantic compression is lossy as structural schemas discard the linguistic context and temporal markers essential for vector retrieval.
This failure incurs a prohibitive cost since \texttt{Rewrite} inflates the insertion latency for Queue+Segment by over \textbf{$10\times$} from 146\,ms to 1552\,ms. 
This inefficiency persists even when scaling to stronger embedding backbones like E5-Large as discussed in Appendix~\ref{app:detailed_experimental_results} which proves that preserving raw texture is far more efficient than structured abstraction.

\noindent\textit{\textbf{Robustness Check with SOTA MoE Model:}} 
We further validated our findings on a representative subset of the LoCoMo dataset using Qwen-Plus model applied to the Property Graph architecture, with a relaxed extraction limit of 10 triplets.
As shown in Table~\ref{tab:robustness_qwen}, Even with this enhanced setup, the \texttt{Rewrite} strategy caused the Mean F1 to fall from 0.389 achieved by the \texttt{None} baseline to 0.163. 
This performance degradation accompanies a latency increase from 111\,ms to 1161\,ms, confirming that structural information loss is a fundamental limitation rather than a symptom of insufficient model capacity.

\begin{table}[h]
    \centering
    \caption{\textbf{Robustness Check with Qwen-Plus.} Even with a massive capability jump and relaxed extraction limits, \texttt{Rewrite} significantly underperforms \texttt{None} while insertion latency.}
    \label{tab:robustness_qwen}
    \resizebox{\linewidth}{!}{
    \begin{tabular}{lccc}
        \toprule
        \textbf{Strategy} & \textbf{Max Triplets} & \textbf{Mean F1} & \textbf{Latency} \\
        \midrule
        Pangu-\texttt{None} & N/A & 0.324 & 99\,ms \\
        Pangu-\texttt{Rewrite} & 5 & 0.132 & 1887\,ms \\
        \midrule
        Qwen-\texttt{None} & N/A & \textbf{0.389} & \textbf{111\,ms} \\
        Qwen-\texttt{Rewrite} & \textbf{10} & 0.163 & 1161\,ms \\
        \bottomrule
    \end{tabular}
    }
    \vspace{-10pt}
\end{table}

\noindent\textbf{O\themycounter\stepcounter{mycounter}: Enrichment is a redundant luxury.}
The \texttt{Enrich} strategy imposes prohibitive costs for negligible gains, exemplified by Queue+Segment where insertion latency quadruples from 146\,ms to 580\,ms without improving F1 scores.
Similarly, Property Graph observes only a transient benefit in the first round that completely evaporates by Round 5, despite the summarization process doubling the write time to over 2.5\,s.
This negative trade-off is visualized in the cost-effectiveness trajectories of Figure~\ref{fig:four_subfigures}.(a), confirming that the heavy computational tax of preprocessing drags down overall system efficiency.
Since modern embeddings effectively capture semantics from raw segments, minimalism on the insertion pipeline remains the optimal strategy for long-horizon deployment.

\subsection{Consolidation Policy}
\label{subsec:consolidation_policy}

Consolidation updates the memory data structures state according to maintenance policies. We instantiate representative algorithms across three paradigms including \texttt{llm\_crud}~\cite{mem0} for Conflict Resolution, \texttt{forgetting\_curve}~\cite{li2025helloagainllmpoweredpersonalized} for Decay Eviction, and the \texttt{heat\_migration}~\cite{kang2025memoryosaiagent} and \texttt{link\_evolution}~\cite{mem0} heuristics for Structure Enrichment. Table~\ref{tab:consolidation_policy} contrasts these interventions against the unmaintained baseline.

\begin{table}[h]
    \centering
    \caption{\textbf{Impact of Consolidation Policies.} We compare the baseline against proactive strategies using latency. The data shows that while \texttt{CRUD} offers marginal accuracy improvements, it imposes a prohibitive post-processing latency penalty compared to the lightweight \texttt{Forgetting} curve.}
    \label{tab:consolidation_policy}
    \resizebox{\linewidth}{!}{
    \begin{tabular}{ll ccc r}
    \toprule
    \multirow{2}{*}{\textbf{System}} & \multirow{2}{*}{\textbf{Strategy}} & \multicolumn{3}{c}{\textbf{Token-level F1}} & \multirow{2}{*}{\textbf{Latency (ms)}} \\
    \cmidrule(lr){3-5}
     & & \textbf{R1} & \textbf{R5} & \textbf{Mean} \\
    \midrule
    \multirow{3}{*}{Property Graph} 
     & \texttt{None} & 0.348 & 0.269 & 0.300 & \textbf{$<1$} \\
     & \texttt{CRUD} & \textbf{0.351} & \textbf{0.273} & 0.301 & 3,777 \\
     & \texttt{Forgetting} & 0.343 & 0.270 & \textbf{0.303} & 120 \\
    \midrule
    \multirow{3}{*}{Queue+Segment} 
     & \texttt{None} & \textbf{0.370} & 0.317 & 0.341 & \textbf{$<1$} \\
     & \texttt{CRUD} & \textbf{0.370} & \textbf{0.319} & \textbf{0.342} & 1,983 \\
     & \texttt{Decay Eviction} & \textbf{0.370} & 0.318 & \textbf{0.342} & 61 \\
    \midrule
    \multirow{3}{*}{LSH Hash} 
     & \texttt{None} & 0.102 & 0.095 & 0.101 & \textbf{$<1$} \\
     & \texttt{CRUD} & \textbf{0.103} & 0.095 & 0.100 & 2,894 \\
     & \texttt{Forgetting} & 0.101 & \textbf{0.096} & \textbf{0.101} & \textbf{$<1$} \\
    \bottomrule
    \end{tabular}
    }
    \vspace{-10pt}
\end{table}

\noindent\textbf{O\themycounter\stepcounter{mycounter}: In online streaming regimes, heuristic consolidation outperforms generative maintenance.}
The comparative analysis within the strict latency constraints of online streaming reveals that while generative strategies like \texttt{CRUD} aim to enhance consistency, they act as a prohibitive bottleneck for real-time pipelines. 
For the Property Graph architecture, enabling LLM-based resolution inflates the post-processing latency to 3,777\,ms while only improving the mean F1 score by a marginal 0.001. 
Such multi-second overheads render active conflict resolution impractical for synchronous user interactions despite potential logical benefits.
In contrast, lightweight heuristics achieve superior efficiency-utility trade-offs suitable for online deployment. 
On the Queue+Segment system, the \texttt{Decay Eviction} policy matches the peak accuracy of the expensive \texttt{CRUD} strategy at 0.342 but operates with a latency of only 61\,ms compared to the 1,983\,ms required for the generative approach. 
This confirms that algorithmic maintenance offers a strictly dominant strategy for online memory systems: it provides the deterministic control necessary for policy-driven forgetting without the prohibitive latency of generative models, effectively distinguishing intentional state management from the inevitable temporal degradation.

\subsection{Query Formulation Strategy}
\label{subsec:query_formulation_strategy}

The Query Formulation dimension serves as the translational interface that converts raw user intent into actionable retrieval signals. 
Beyond the standard \texttt{None} baseline, we examine three distinct paradigms: the \texttt{Validate} heuristic for query gating, \texttt{Keyword} extraction for optimization, and the \texttt{Decompose} method for generative enhancement. 
Table~\ref{tab:query_formulation} reports the resulting streaming performance, specifically contrasting the query formulation overhead with architectural insertion costs.

\begin{table}[h]
    \centering
    \caption{\textbf{Performance and Latency of Query Formulation Strategies.} Generative strategies incur prohibitive latency without yielding proportional accuracy gains over the baseline.}
    \label{tab:query_formulation}
    \resizebox{\linewidth}{!}{
    \begin{tabular}{ll ccc rr}
    \toprule
    \multirow{2}{*}{\textbf{System}} & \multirow{2}{*}{\textbf{Strategy}} & \multicolumn{3}{c}{\textbf{Token-level F1}} & \multicolumn{2}{c}{\textbf{Latency (ms)}} \\
    \cmidrule(lr){3-5} \cmidrule(lr){6-7} & & \textbf{R1} & \textbf{R5} & \textbf{Mean} & \textbf{PreRet.} & \textbf{Ins.} \\
    \midrule
    \multirow{3}{*}{Property Graph} & \texttt{None} & \textbf{0.342} & \textbf{0.273} & \textbf{0.300} & 100 & 5742 \\
    & \texttt{Keyword} & 0.102 & 0.082 & 0.091 & 870 & 5394 \\ 
    & \texttt{Decompose} & 0.320 & 0.246 & 0.279 & 1913 & 5182 \\
    \midrule
    \multirow{3}{*}{Queue+Segment}
    & \texttt{None} & 0.371 & \textbf{0.318} & \textbf{0.341} & 161 & 187 \\
    & \texttt{Keyword} & 0.114 & 0.083 & 0.099 & 1117 & 193 \\
    & \texttt{Decompose} & 0.367 & 0.302 & 0.328 & 2209 & 209 \\
    \midrule
    \multirow{3}{*}{LSH Hash}
    & \texttt{None} & 0.065 & 0.061 & 0.063 & 157 & 1963 \\
    & \texttt{Keyword} & 0.030 & 0.030 & 0.031 & 1064 & 2056 \\
    & \texttt{Decompose} & \textbf{0.093} & \textbf{0.090} 
    & \textbf{0.094} & 2352 & 2016 \\
    \bottomrule
    \end{tabular}
    }
    \vspace{-10pt}
\end{table}

\noindent\textbf{O\themycounter\stepcounter{mycounter}: Generative query formulation acts as a latency trap with negative returns.}
The data in Table~\ref{tab:query_formulation} indicates that deploying LLM-based strategies like \texttt{Keyword} extraction or \texttt{Decompose} often introduces efficiency bottlenecks that outweigh their benefits compared to direct processing.
For instance, reducing queries to keywords leads to significant performance degradation in Queue+Segment, where the Mean F1 score drops from 0.341 using the \texttt{None} baseline to 0.099 with \texttt{Keyword}.
This decline in accuracy is accompanied by a substantial latency increase, as the generation step adds between 870\,ms and 1117\,ms to the pipeline, contrasting sharply with standard vector lookups that take roughly 160\,ms.
Similarly, the \texttt{Decompose} strategy imposes a heavy computational tax while degrading performance.
While Queue+Segment demonstrates inherent efficiency with insertion and retrieval latencies under 200\,ms, enabling decomposition inflates the retrieval time by over ten times to 2209\,ms while resulting in a slightly lower F1 score of 0.328.
This suggests that for robust storage backbones, preserving semantic context via direct retrieval often offers a more favorable cost-benefit ratio than expensive generative interventions.


\subsection{Context Integration Mechanism}
\label{subsec:context_integration_mechanism}

Our analysis concludes with the Context Integration Mechanism, which acts as the interface bridging discrete retrieval results and the continuous generation process. 
We compare the lightweight heuristic \texttt{Augment} against the generative expansion strategy \texttt{Multi-query}, with detailed breakdowns provided in Table~\ref{tab:context_integration_main} and Table~\ref{tab:context_integration_llama}.

\begin{table}[h]
\centering
\caption{\textbf{Impact of Context Integration.} Generative fusion incurs a massive latency tax compared to heuristics. It spikes context integration latency by orders of magnitude without delivering meaningful F1 gains.}
\label{tab:context_integration_main}
\resizebox{\linewidth}{!}{
\begin{tabular}{ll ccccc cc}
\toprule
\multirow{2}{*}{\textbf{System}} & \multirow{2}{*}{\textbf{Strategy}} & \multicolumn{5}{c}{\textbf{Token-level F1 per Round}} & \multirow{2}{*}{\textbf{Mean F1}} & \multirow{2}{*}{\textbf{Latency}} \\
\cmidrule(lr){3-7} 
 & & \textbf{R1} & \textbf{R2} & \textbf{R3} & \textbf{R4} & \textbf{R5}  \\
\midrule
\multirow{3}{*}{Property Graph} 
 & \texttt{None} & \textbf{0.346} & 0.301 & \textbf{0.294} & 0.286 & 0.271 & \textbf{0.299} & \textbf{$<1$} \\
 & \texttt{Augment} & 0.333 & \textbf{0.309} & 0.290 & \textbf{0.288} & 0.270 & 0.298 & $<1$ \\
 & \texttt{Multi-query} & 0.335 & 0.305 & 0.293 & 0.286 & \textbf{0.274} & 0.299 & 1700 \\
\midrule
\multirow{3}{*}{LSH Hash} 
 & \texttt{None} & 0.095 & \textbf{0.096} & \textbf{0.101} & \textbf{0.100} & \textbf{0.094} & \textbf{0.097} & \textbf{$<1$} \\
 & \texttt{Augment} & 0.097 & 0.094 & 0.087 & 0.083 & 0.076 & 0.088 & $<1$ \\
 & \texttt{Multi-query} & \textbf{0.097} & 0.089 & 0.096 & 0.093 & 0.083 & 0.092 & 1584 \\
\bottomrule
\end{tabular}
}
\vspace{-10pt}
\end{table}

\noindent\textbf{O\themycounter\stepcounter{mycounter}: Multi-query expansion acts as a ``latency tax'' with diminishing returns.}
While generative fusion is often hypothesized to improve robustness, our results in Table~\ref{tab:context_integration_main} define it strictly as a \textbf{latency tax}.
For Property Graph, enabling \texttt{Multi-query} yields zero improvement in Mean F1, stagnating at 0.299, yet it explodes post-processing latency from negligible levels below 1\,ms to approximately 1.7\,s.
The trend worsens for LSH Hash, where performance actually drops from 0.097 to 0.092 despite the same massive latency penalty.
This confirms that, at least for the default backbone, the cost is paid in seconds while the value is either non-existent or negative.

\begin{table}[h]
    \centering
    \caption{\textbf{Robustness Verification with Llama-3-8B.} Although \texttt{Multi-query} slightly outperforms \texttt{Augment}, it incurs a prohibitive latency cost compared to the heuristic baseline.}
    \label{tab:context_integration_llama}
    \resizebox{\linewidth}{!}{
    \begin{tabular}{l ccccc cc}
    \toprule
    \multirow{2}{*}{\textbf{Strategy}} & \multicolumn{5}{c}{\textbf{Token-level F1 per Round}} & \multirow{2}{*}{\textbf{Mean F1}} & \multirow{2}{*}{\textbf{Latency}} \\
    \cmidrule(lr){2-6} 
     & \textbf{R1} & \textbf{R2} & \textbf{R3} & \textbf{R4} & \textbf{R5} \\
    \midrule
    \texttt{Augment} & 0.339 & 0.301 & 0.289 & 0.283 & 0.268 & 0.296 & \textbf{$<1$} \\
    \texttt{Multi-query} & \textbf{0.348} & \textbf{0.306} & \textbf{0.295} & \textbf{0.289} & \textbf{0.279} & \textbf{0.303} & 1327 \\
    \bottomrule
    \end{tabular}
    }
    \vspace{-10pt}
\end{table}

\noindent\textit{\textbf{Robustness Check with Llama-3-8B:}}
As detailed in Table~\ref{tab:context_integration_llama}, switching to the stronger Llama-3-8B backbone reveals a nuanced trade-off. 
While \texttt{Multi-query} successfully outperforms \texttt{Augment} by raising the Mean F1 from 0.296 to 0.303, this accuracy gain comes at a steep price. 
The integration latency surges from negligible levels ($<1$\,ms) to 1327\,ms. 
This suggests that while stronger models can extract value from generative fusion, placing these steps on the retrieval critical path remains a costly strategy, likely suitable only when precision is paramount and latency is not a primary constraint.


\subsection{Limitation}
\label{subsec:limitation}

The scope of our analysis is primarily bounded by the computational intractability of iterative graph refinement where paradigms like \hipporag were excluded due to optimization latencies that scale poorly with memory size. 
Such overheads can extend execution times from minutes to hours effectively rendering these systems incompatible with the strict timeouts of real-time conversational pipelines. 
This constraint on algorithmic coverage is paralleled by an empirical limitation where the scarcity of datasets explicitly curated for dynamic memory evolution precludes the granular monitoring of intermediate states. 
Consequently our evaluation relies on end-to-end outcome metrics rather than precise state verification. 
Moreover, the absence of established definitions distinguishing short-term working memory from long-horizon storage prevents the application of hybrid consolidation strategies that could otherwise optimize memory state updates based on retention scope, identifying the formalization of these hierarchical tiers as a critical direction for future architectural research.

\compact

\section{Conclusion}
\label{sec:conclusion}

This paper investigates \memory in the realistic streaming regime where insertions interleave with retrievals. 
We introduce \system to decompose the lifecycle into five dimensions, enabling granular attribution of answer quality and system latency. 
Our results reveal that the storage structure largely dictates the accuracy ceiling, while content-destructive transformations often degrade performance despite their overhead. 
Crucially, we identify a distinct ``latency tax'' in generative optimizations: while stronger backbones can extract utility from strategies like multi-query expansion, the computational cost remains prohibitive for real-time deployment compared to lightweight heuristics. 
Furthermore, we demonstrate that heuristic maintenance offers a superior trade-off for online systems, providing the deterministic control necessary for privacy compliance without the latency of generative approaches. 
\system establishes a reusable substrate for future work to optimize the full insert--maintain--retrieval--integrate loop against the inevitable entropy of long-horizon interaction.
\compact


\compact


\compact
\section*{Impact Statement}

This work aims to advance the field of Machine Learning by focusing on the evaluation and optimization of External Memory Modules for Large Language Models under realistic streaming conditions.

From a societal and environmental perspective, our granular decomposition of the memory lifecycle identifies a critical ``latency tax'' inherent in current generative designs. We demonstrate that aggressive semantic compression and post-retrieval fusion often incur prohibitive computational costs for negligible or even negative performance gains. By establishing that lightweight structural heuristics can anchor the efficiency frontier, our findings offer a rigorous pathway toward more energy-efficient and scalable AI systems, directly supporting the principles of Green AI.

From an ethical and privacy perspective, the shift to a streaming protocol highlights the risks of unbounded data accumulation. 
Our analysis confirms that heuristic consolidation policies, such as decay eviction strategies, offer not only superior latency profiles but also the deterministic control necessary for privacy compliance. 
Unlike non-deterministic generative maintenance, these mechanisms provide a reliable foundation for adhering to data erasure mandates and preventing the permanent retention of sensitive user data in long-horizon personalized agents.
We advocate for future research to prioritize such privacy-preserving memory updates alongside accuracy and latency optimization.

\bibliographystyle{ACM-Reference-Format}
\bibliography{References}

\newpage
\appendix
\onecolumn
\subcompact



\section{Supplement}
\label{sec:supplement}


\subsection{Related Work}
\label{app:related_work}

\subsubsection{External Memory Modules}
\label{app:external_memory_modules}
We analyze twelve representative \memory along the five lifecycle stages (D1–D5) defined in Section~\ref{sec:design_of_EMM}. Consistent with our taxonomy, we categorize these systems into two architectural paradigms based on their topological organization: \textit{Hierarchical and Graph-Structured Services} and \textit{Partitional Memory Services}.

\textbf{Hierarchical and Graph-Structured Modules.}
Systems such as \hipporag (and HippoRAG 2)~\cite{gutierrez2025hipporag}, \memzerog~\cite{mem0}, and \amem~\cite{xu2025amemagenticmemoryllm} align with the hierarchical paradigm by explicitly modeling the topological relationships between memory units. Rather than treating memories as independent vectors, these architectures construct semantic knowledge graphs (D1) where edges encode relational dependencies. This necessitates rigorous extraction-based normalization (D2) to transform raw text into triples or linked entities. Consequently, consolidation (D3) becomes a structural operation involving link evolution or graph densification, while integration (D5) leverages graph traversal algorithms to retrieve multi-hop context unreachable via simple similarity search.

\textbf{Partitional Memory Modules.}
The remaining systems employ partitional architectures (D1), treating memory as a flat pool managed through composed indices (e.g., vector stores, temporal queues, keyword tables). Rather than modeling explicit topology, these services optimize system behavior through the strategic composition of retrieval structures. One prominent direction, adopted by \memoryos~\cite{kang2025memoryosaiagent}, \memgpt~\cite{packer2024memgptllmsoperatingsystems}, \scm~\cite{wang2024enhancing}, \ldagent~\cite{li2025helloagainllmpoweredpersonalized}, and \memorybank~\cite{zhong2023memorybankenhancinglargelanguage}, leverages this flexibility for \textit{lifecycle and capacity management}. By organizing memory into temporal tiers (e.g., short-term queues vs. long-horizon storage), these systems implement dynamic consolidation policies (D3)—such as recursive summarization or operating-system-inspired paging—to handle infinite context streams within bounded resource constraints. Conversely, systems like \memzero~\cite{mem0}, \tim~\cite{liu2023thinkinmemoryrecallingpostthinkingenable}, and \secom~\cite{pan2025memoryconstructionretrievalpersonalized} utilize partitional indexing to enhance \textit{semantic precision and consistency}. These approaches discretize inputs into independent atomic units (D2) and employ granular indexing strategies (D4) to support targeted updates. This design facilitates rigorous conflict resolution (D3), enabling the system to detect and correct contradictions via LLM-driven operations efficiently, a task often computationally prohibitive in dense graph structures.

\subsubsection{Memory Benchmark}
\label{app:memory_benchmark}

We categorize existing benchmarks based on their evaluation protocols and measurement focus, contrasting them with our streaming approach. Table~\ref{tab:benchmark-comparison} summarizes these dimensions.

Most established benchmarks operate under an \textbf{offline protocol}, decoupling memory construction from querying. \locomo~\cite{locomo} targets long-horizon information synthesis in dyadic dialogues, while \longmemeval~\cite{longbench} focuses on recall consistency across multi-session chats. Both process static logs where memory is built in a single pass. \membench~\cite{membench} similarly evaluates retrieval from a frozen memory pool. An exception is \hustmem~\cite{hustbench}, which supports incremental ingestion for agentic tasks (e.g., fact consolidation); however, it typically defers queries to the end of the session, failing to capture the complexity of fully interleaved production-retrieval cycles.

Existing benchmarks also vary significantly in evaluation granularity and measurement coverage. \locomo evaluates at the \textbf{model level} (end-to-end output), whereas others like \membench and \longmemeval focus on the \textbf{memory module's} isolation. Crucially, measurement often prioritizes accuracy over system cost. \locomo, \longmemeval, and \hustmem report only performance metrics (e.g., accuracy, consistency). While \membench includes simple efficiency metrics such as Read Time (RT) and Write Time (WT), it does not distinguish between model inference and memory maintenance costs, nor does it track efficiency evolution over time.

\textbf{Our Work} bridges these gaps by proposing a testbed at the \textbf{operator level} with a \textbf{streaming protocol}. By interleaving insertion and retrieval, we evaluate performance on evolving memory states $M^{(k_t)}$. Unlike prior works, we provide a dual-track assessment of both accuracy (F1) and comprehensive efficiency (insertion/retrieval latency, memory footprint), explicitly decoupling memory system overhead from the underlying LLM.

\begin{table*}[h]
\centering
\caption{Comparison of memory benchmarks across evaluation granularity, protocol, and measurement coverage.}
\label{tab:benchmark-comparison}
\small
\setlength{\tabcolsep}{6pt}
\begin{tabularx}{\textwidth}{l>{\raggedright\arraybackslash}X>{\raggedright\arraybackslash}X>{\raggedright\arraybackslash}X}
\toprule
\textbf{Benchmark} & \textbf{Evaluation Granularity} & \textbf{Evaluation Protocol} & \textbf{Measurement Coverage} \\
\midrule
LoCoMo & Model level & Offline & Accuracy only \\
\addlinespace
MemoryAgentBench & Module level & Incremental Insertion & Accuracy only \\
\addlinespace
LongMemEval & Module level & Offline & Accuracy only \\
\addlinespace
MemBench & Module level & Offline & Accuracy + Partial efficiency \\
\midrule
\textbf{Our Work} & \textbf{\textit{Operator level}} & \textbf{\textit{Interleaved Insertion/Retrieval}} & \textbf{\textit{Accuracy + Efficiency}} \\
\bottomrule
\end{tabularx}
\end{table*}


\subsection{Detailed Design Aspects of External Memory Modules}
\label{app:design_aspects_of_EEMs}
We deconstruct the twelve baseline systems evaluated in this study according to the five lifecycle stages (D1--D5). For each stage, we identify the primary design paradigms and categorize the baselines accordingly.

\subsubsection{D1: Memory Data Structure}
We classify memory architectures into two dominant paradigms based on their topological organization (D1). \texttt{Partitional} services treat memory as a flat pool and rely on multiple complementary indices to support diverse access patterns. For instance, \memoryos combines FIFO queues with segmented storage to manage temporal tiers, while \memorybank utilizes a combination of vector stores and summary buffers. This approach excels at predictable latency and scalability but may struggle with deep relational queries. In contrast, \texttt{Hierarchical} services embed memories into explicit relational structures. \memzerog, for example, constructs semantic knowledge graphs where nodes represent entities and edges encode dependencies, enabling multi-hop retrieval beyond nearest-neighbor similarity. While offering richer reasoning primitives, these structures introduce additional maintenance complexity. Table~\ref{tab:d1_categorization} maps the evaluated systems to these structural paradigms.

\begin{table}[H]
    \centering
    \caption{Summary of Memory Data Structures (D1) in representative models.}
    \label{tab:d1_categorization}
    \small
    \begin{tabularx}{\linewidth}{XXl}
        \toprule
        \textbf{Structure Type} & \textbf{Representative Models} & \textbf{Specific Implementation} \\
        \midrule
        Hierarchical/Graph &\memzerog & \path{semantic_inverted_knowledge_graph} \\
        \midrule
        Partitional        & \memzero & \path{inverted_vectorstore_combination} \\
                           & \tim & \texttt{lsh\_hash} \\
                           &  \memoryos & \path{feature_queue_segment_combination} \\
                           & \memorybank & \path{feature_summary_vectorstore_combination} \\
        \bottomrule
    \end{tabularx}
\end{table}

\subsubsection{D2: Normalization Strategy}
Normalization strategies (D2) transform raw conversational inputs into storable units to balance \emph{fidelity} with \emph{indexability}. Strategies generally fall into three categories: \texttt{Direct Storage} (identity mapping), employed by models like \memoryos to maximize fidelity; \texttt{Enrichment}, which augments input via summarization or tagging to improve index compatibility, as seen in \memorybank where raw interactions are summarized into events; and \texttt{Rewriting}, which converts unstructured dialogue into structured formats. For example, \tim and \memzerog extract relational triplets or atomic facts to support structured retrieval. While rewriting enhances relational reasoning, it increases dependence on upstream model quality compared to simpler enrichment or direct storage. Table~\ref{tab:d2_categorization} categorizes the baseline models by their specific normalization operators.

\begin{table}[H]
    \centering
    \caption{Summary of Normalization Strategies (D2) in representative models.}
    \label{tab:d2_categorization}
    \small
    \begin{tabularx}{\linewidth}{XXX}
        \toprule
        \textbf{Strategy} & \textbf{Representative Models} & \textbf{Specific Operator} \\
        \midrule
        None/Embedding 
                   & \memoryos & \texttt{none} \\
        \midrule
        Enrich 
                   & \memorybank & \texttt{summarize} \\
        \midrule
        Rewrite    & \tim, \memzerog & \texttt{triplet\_extract} \\
        \bottomrule
    \end{tabularx}
\end{table}

\subsubsection{D3: Consolidation Policy}
Consolidation policies (D3) manage the lifecycle of stored memories. We observe three main families: \texttt{Conflict Resolution} handles redundancy and contradictions; \memzerog implements this via LLM-driven CRUD operations to maintain consistency. \texttt{Decay Eviction} manages capacity constraints; \memorybank uses an Ebbinghaus-inspired forgetting curve, while \memoryos employs heat-based migration to move data between tiers. \texttt{Structure Enrichment} evolves the relational topology, exemplified by \amem, which dynamically creates links between memory nodes. These policies improve long-horizon memory quality but often incur maintenance costs. Table~\ref{tab:d3_categorization} details the specific consolidation policies employed by each representative model.

\begin{table}[H]
    \centering
    \caption{Summary of Consolidation Policies (D3) in representative models.}
    \label{tab:d3_categorization}
    \small
    \begin{tabularx}{\linewidth}{XXX}
        \toprule
        \textbf{Policy} & \textbf{Representative Models} & \textbf{Specific Operator} \\
        \midrule
        None                & \scm & \texttt{none} \\
        \midrule
        Conflict Resolution & \memzerog & \texttt{llm\_crud} \\
                            & \tim & \texttt{semantic\_consolidation} \\
        \midrule
        Decay Eviction      & \memorybank& \texttt{forgetting\_curve} \\
                            & \memoryos & \texttt{heat\_migration} \\
        \midrule
        Structure Enrichment& \amem & \texttt{link\_evolution} \\
        \bottomrule
    \end{tabularx}
\end{table}

\subsubsection{D4: Query Formulation Strategy} 
Query Formulation Strategies (D4) address the mismatch between user intentions and memory structures. Beyond \texttt{Direct Processing} (used by \memzerog, \memorybank, and \tim), systems employ more complex strategies. \texttt{Query Enhancement} and \texttt{Validation} can be used to decompose complex questions or verify retrieval necessity. \texttt{Query Optimization} is adopted by \memoryos, which extracts keywords from the query to target specific memory indices. These strategies transform the user’s original query into a form better suited to the underlying memory structure, improving accuracy at the cost of latency. Table~\ref{tab:d4_categorization} lists the query formulation strategies adopted across the evaluated systems.

\begin{table}[H]
    \centering
    \caption{Summary of Query Formulation Strategies (D4) in representative models.}
    \label{tab:d4_categorization}
    \small
    \begin{tabularx}{\linewidth}{XXX}
        \toprule
        \textbf{Strategy} & \textbf{Representative Models} & \textbf{Specific Operator} \\
        \midrule
        None/Embedding & \memzerog, \memorybank, \tim& \texttt{embedding} \\
        \midrule
        Validate         & - & \texttt{validate} \\
        \midrule
        Optimization     & \memoryos & \texttt{keyword\_extract} \\
        \midrule
        Enhancement      & - & \texttt{decompose} \\
        \bottomrule
    \end{tabularx}
\end{table}

\subsubsection{D5: Context Integration Mechanism}
Context Integration Mechanisms (D5) determine what is presented to the generator model. Common strategies include \texttt{Filtering}, where systems like \memzero use similarity thresholds to reduce noise; and \texttt{Reranking}, where \memorybank uses time-weighted scoring to prioritize relevance. \texttt{Merging} strategies are critical in multi-tiered systems like \memoryos, which aggregates results from different storage layers. This stage manages the trade-off between \emph{context quality} and \emph{cost}: aggressive filtering reduces noise but risks harming recall, while complex reranking improves relevance at the expense of latency. Table~\ref{tab:d5_categorization} summarizes the context integration mechanisms implemented in the baseline models.

\begin{table}[H]
    \centering
    \caption{Summary of Context Integration Mechanisms (D5) in representative models.}
    \label{tab:d5_categorization}
    \small
    \begin{tabularx}{\linewidth}{XXX}
        \toprule
        \textbf{Strategy} & \textbf{Representative Models} & \textbf{Specific Operator} \\
        \midrule
        None   & \memzerog, \tim & \texttt{none} \\
        \midrule
        Rerank & \memorybank & \texttt{time\_weighted} \\
        \midrule
        Filter &\memzero& \texttt{threshold} \\
        \midrule
        Merge  & \memoryos & \texttt{multi\_tier} \\
        \bottomrule
    \end{tabularx}
\end{table}




\subsection{Dataset Details and Preprocessing}
\label{app:dataset_details}
We evaluate \memory on a suite of specialized datasets, adapting each to the \emph{streaming} protocol to quantify the evolution of memory quality and cost over time.

\textbf{LoCoMo.} The LoCoMo dataset is structured around 10 core tasks, where each task comprises multiple sessions containing multi-turn dialogue sequences. Crucially, each task is associated with a specific set of tests accompanied by evidence labels that pinpoint the exact dialogue history required for reasoning. This hierarchical organization makes LoCoMo inherently suitable for adaptation into a streaming format. In alignment with our streaming protocol, we serialize these multi-session dialogues into a continuous chronological stream, where evaluation queries are triggered immediately after their supporting evidence is ingested, ensuring strict temporal causality.

\textbf{LongMemEval.} To assess system stability over extended interaction periods, we utilize the \texttt{LongMemEvalOracle} subset of \longmemeval. The original benchmark structures each task as a sequence of $N$ historical turns followed by a single query. We adapt this for streaming evaluation by concatenating multiple independent tasks into a unified chronological stream. This approach creates a long-horizon timeline where questions are triggered at specific intervals, validating the system's ability to maintain and recall context across disjointed interaction sessions.

\textbf{MemoryAgentBench.} To evaluate the handling of evolving information, we incorporate the \texttt{Selective Forgetting} subset from MemoryAgentBench. Our analysis of the full benchmark revealed a significant presence of static factual knowledge; consequently, we exclusively selected the \texttt{Selective Forgetting} subset to construct our dataset, ensuring a focus on dynamic consistency. This subset simulates real-world scenarios where knowledge changes over time, requiring the memory module to correctly consolidate up-to-date facts from an incremental stream while effectively disregarding obsolete data.

To support our streaming evaluation, we parse the sequential fact updates into a discrete input stream and align the corresponding questions and ground-truth answers as synchronized evaluation checkpoints, ensuring that the memory system is tested on its real-time capacity to integrate new information and resolve contradictions.


\subsection{Algorithmic Details of Memory Operations}
\label{app:algorithms_fg2}

We provide pseudocode for the two atomic operations of \memoryone, aligning with Equations~\eqref{eq:insert} and~\eqref{eq:retrieve}.

\subsubsection{Memory State Update}

Triggered by an \textsc{Insert} request, this operation updates the memory state from $M^{(k-1)}$ to $M^{(k)}$ via preprocessing, insertion, and optimization.

\begin{algorithm}[!h]
    \caption{Algorithm of Memory Update Step (\textsc{Insert})}
    \label{alg:update}
    
    \begin{algorithmic}[1]
        \REQUIRE Current memory state $M^{(k-1)}$, Historical context $h^{(k)}$  
        \ENSURE Updated memory state $M^{(k)}$
        \HYPERPARAMETERS Preprocessing config $\phi$, Optimization threshold $\theta$
        \STATE Step 1: Context Preprocessing
        \STATE \quad $\tilde{h} \gets \textsc{PreIns}(h^{(k)}; \phi)$ 

        \STATE Step 2: Tentative Insertion
        \STATE \quad $M_{\text{inter}} \gets \textsc{Insert}(M^{(k-1)}, \tilde{h})$

        \STATE Step 3: Global Optimization (e.g., pruning, merging)
        \STATE \quad $M^{(k)} \gets \textsc{Optimize}(M_{\text{inter}}; \theta)$

        \STATE \textbf{return} $M^{(k)}$
    \end{algorithmic}
\end{algorithm}

\subsubsection{Context Retrieval}

Given a query $q$ at time $\tau$, retrieval accesses the latest memory state $M^{(k^*)}$ (where $k^*$ is the number of \textsc{Insert} requests with $\tau_i \leq \tau$) and returns context $c$.

\begin{algorithm}[h]
   \caption{Context Retrieval Step (\textsc{Retrieve})}
   \label{alg:retrieve}

\begin{algorithmic}[1]
   \REQUIRE Current memory state $M^{(k^*)}$, Raw user query $q$
   \ENSURE Retrieved context $c$
   \HYPERPARAMETERS Query refinement config $\psi$, Top-$k$ parameter $K$
   \STATE{Step 1: Query Refinement}
   \STATE \quad$\tilde{q} \gets \textsc{PreQry}(q; \psi)$
   \STATE{Step 2: Similarity Search \& Extraction}
   \STATE \quad$\mathcal{C}_{\text{cand}} \gets \textsc{Search}(M^{(k^*)}, \tilde{q}, K)$
   \STATE{Step 3: Context Aggregation}
   \STATE \quad$c \gets \textsc{RefineRetrieve}(\mathcal{C}_{\text{cand}})$
   
   \STATE \textbf{return} $c$
\end{algorithmic}
\end{algorithm}

\subcompact
\section{Implementation Details}
\label{app:implementation}

This appendix provides a comprehensive technical breakdown of the \textsc{Neuromem} framework to ensure the reproducibility and clarity of our experimental protocol. We first detail the engineering implementation of the interleaved insertion-and-retrieval protocol, focusing on the pipeline orchestration and backpressure mechanisms used to enforce temporal causality. Subsequently, we specify the hybrid hardware and software infrastructure utilized for different model backbones. Finally, we outline the standardized experimental workflow and the roadmap for open-source deployment.

\subsection{Implementation of Interleaved Insertion-and-Retrieval Protocol}
\label{sec:appendix_implementation}

\begin{figure}[h]
    \centering
    \includegraphics[width=0.9\linewidth]{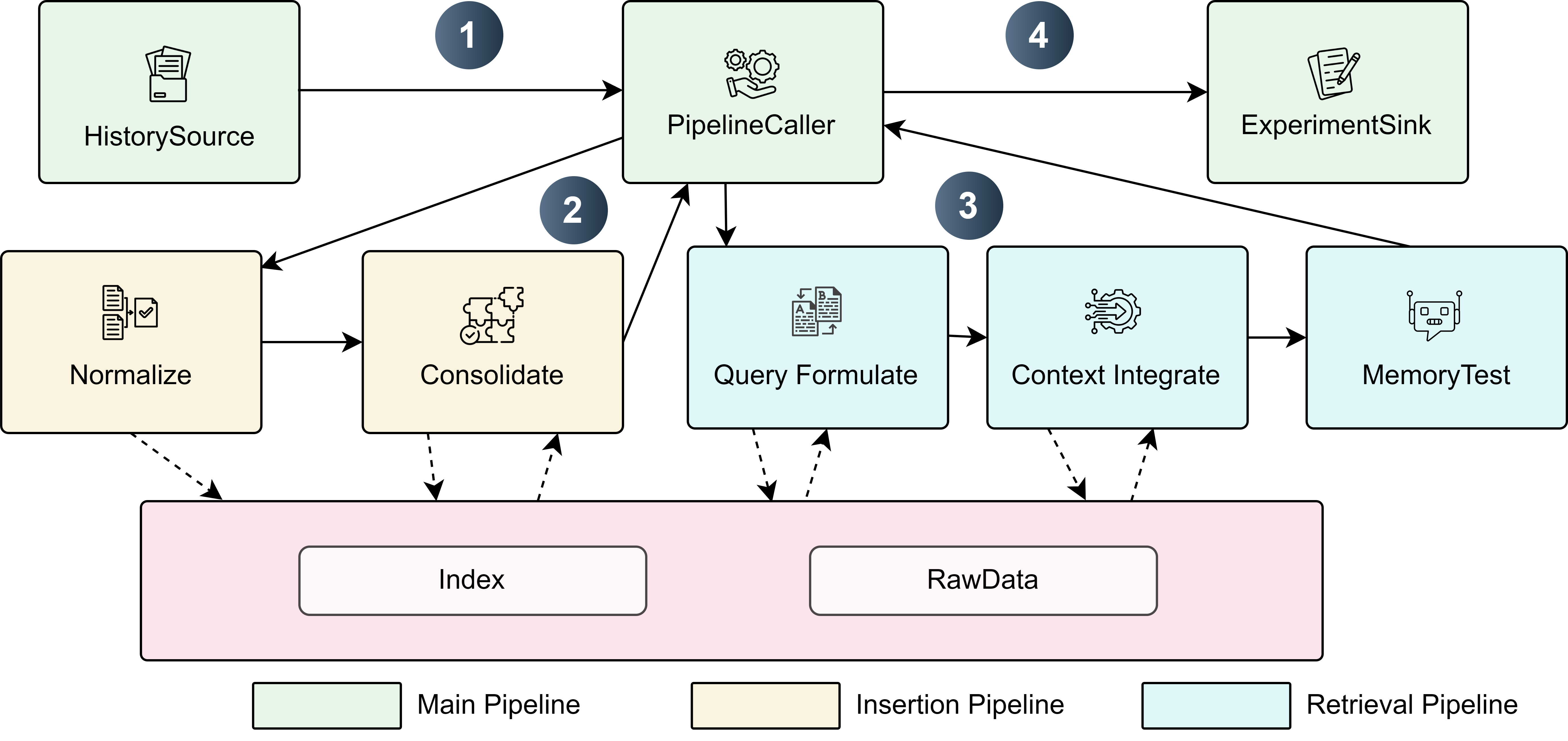}
    \caption{\textbf{Overview of the Neuromem Streaming Pipeline Architecture.} The central \texttt{PipelineCaller} holds the test dataset and orchestrates the interleaved execution. It employs a backpressure mechanism to block the historical data stream during memory maintenance or reasoning evaluation, ensuring strict temporal causality.}
    \label{fig:pipeline_arch}
\end{figure}

To support the streaming evaluation protocol defined in Section 4.1, we implemented a pipeline-based architecture in \textsc{Neuromem} that strictly enforces the causal ordering of the request stream $R$. As illustrated in Figure~\ref{fig:pipeline_arch}, the architecture comprises a central coordinator designated as the Main Pipeline, along with two functional sub-pipelines, namely the Insertion Pipeline and the Retrieval Pipeline, which jointly manage the full lifecycle of the External Memory Module.

\subsubsection{Main Pipeline: The Orchestrator}
The Main Pipeline serves as the backbone of the testbed, governing the flow of time and data. It operates via a strict \textit{backpressure} mechanism to ensure atomicity. It consists of three core components:

\begin{itemize}
    \item \textbf{HistorySource:} Acting as the primary data source, this component is responsible for streaming interaction history data, such as context and facts, in a sequential manner. Unlike static batch loaders, it functions as a stream emitter. Driven by the backpressure mechanism from the downstream caller, the source pauses data emission whenever the processing pipeline is busy, which causes historical data to accumulate in a queue-based buffer rather than overflowing the system.
    
    \item \textbf{PipelineCaller:} Serving as the central coordinator, this component acts as the core scheduler to manage control flow and maintain the evaluation dataset comprising questions and ground truth answers. Rather than passively forwarding data, it actively monitors the incoming stream from the \texttt{HistorySource}. The logic follows a strictly blocking execution cycle that verifies specific state thresholds before triggering downstream processes.
    \begin{enumerate}
        \item \textbf{Insertion Trigger:} Upon receiving each incoming history item $h_t$, the coordinator blocks the stream and synchronously triggers the \textit{Insertion Pipeline} to update the memory state.
        \item \textbf{Threshold Check \& Backpressure:} Following each insertion, the system verifies whether the current state satisfies preset evaluation checkpoints, such as reaching a segment boundary in \textsc{LoCoMo} or achieving a specific fact update count in the \hustmem dataset.
        \item \textbf{Incremental Testing:} If a threshold is met, the \texttt{PipelineCaller} maintains backpressure on the \texttt{HistorySource} by keeping the historical data stream blocked while it initiates the \textit{Retrieval Pipeline} to perform an incremental test. The stream remains blocked until the testing concludes, at which point the system proceeds to process the next historical item $h_{t+1}$.
    \end{enumerate}

    \item \textbf{ExperimentSink:} Functioning as the primary result collector, this component aggregates evaluation metrics, including Token-level F1 scores and latency statistics, from the testing process and serializes them for subsequent analysis.
\end{itemize}

\subsubsection{Insertion Pipeline: State Evolution (D1-D3)}
When activated by the \texttt{PipelineCaller} for a specific history item, this pipeline instantiates Equation (1), $M^{(k)} = \text{POSTINS}(M^{(k-1)}, \text{PREINS}(h^{(k)}))$, into three consecutive stages. The mainstream is blocked until these stages complete:

\begin{enumerate}
    \item \textbf{Normalize:} Implementing the Normalization Strategy, this stage standardizes the raw context using the \texttt{PreInsert} operator. Depending on the configuration, the process ranges from simple text cleaning to complex semantic compression, such as extracting triplets via \texttt{extract.triple} or generating summaries using \texttt{transform.summarize}.
    \item \textbf{State Update:} This stage executes the write operation defined by the Memory Data Structure. The processed data units are committed to the underlying storage substrates, such as an Index or RawData container, by invoking specific service interfaces like \texttt{fifo\_queue.insert()} to finalize physical storage.
    \item \textbf{Consolidate:} Corresponding to the Consolidation Policy, this stage maintains the long-horizon stability of the memory state following the write operation. Key maintenance tasks include resolving conflicts between new and old knowledge via \texttt{conflict\_resolution.llm\_crud} or removing obsolete information by applying policies such as \texttt{decay\_eviction.forgetting\_curve}.
\end{enumerate}

\begin{figure}[h]
    \centering
    \includegraphics[width=0.6\linewidth]{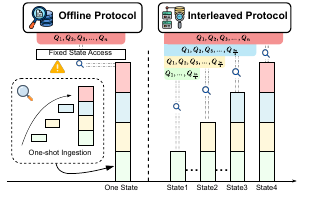}
    \caption{\textbf{Comparison of Evaluation Protocols.} Unlike the traditional \textbf{Offline Protocol} which queries a fixed memory state constructed in a single batch, our \textbf{Interleaved Protocol} performs retrieval at evolving memory states ($State 1, State 2, \dots$), enforcing strict temporal causality.}
    \label{fig:streaming_protocol}
\end{figure}

\subsubsection{Retrieval Pipeline: Reasoning and Evaluation (D1, D4-D5)}

When the \texttt{PipelineCaller} determines that a test checkpoint is reached, it effectively freezes the memory state by blocking new insertions and initiates the Retrieval Pipeline. This mechanism implements the \textbf{Interleaved Protocol} illustrated in Figure~\ref{fig:streaming_protocol} (Right), enabling assessments across dynamically evolving memory states rather than relying on a single fixed snapshot. The process instantiates Equation (2), $c = \text{POSTRET}(M^{(k^*)}, \text{PRERET}(q))$, through the following sequential stages:

\begin{enumerate}
    \item \textbf{Query Formulate:} Implementing the Query Formulation Strategy, this stage utilizes the \texttt{PreRetrieval} operator to convert the user query held by the Caller into executable retrieval signals. Common operations aim to bridge semantic gaps through methods such as \texttt{keyword\_extract} or query rewriting via \texttt{optimize.rewrite}.
    \item \textbf{Context Retrieval:} This stage executes the lookup operation defined by the Memory Data Structure. Using the transformed query signals, the system fetches candidate evidence from the underlying index or raw data storage via specific interfaces like \texttt{fifo\_queue.retrieve()}.
    \item \textbf{Context Integrate:} Corresponding to the Context Integration Mechanism, this stage employs the \texttt{PostRetrieval} operator to refine the fetched candidates. The process synthesizes the final context $c$ for generation by applying filtering or ranking mechanisms, such as \texttt{rerank.time\_weighted} or \texttt{filter.threshold}.
    \item \textbf{Memory Evaluation:} In the final step, the synthesized context is fed into the LLM to generate a response. The system immediately computes evaluation metrics, such as Token-level F1 scores, by comparing the prediction against the ground truth, and transmits the results back to the Main Pipeline.
\end{enumerate}

By employing this strictly blocking Interleaved Protocol, \system ensures that retrieval operations occur on a stable snapshot of the memory state $M^{(k)}$ without interference from incoming data, accurately simulating the real-time constraints of streaming applications.


\subsection{Hardware and Software Infrastructure}
\label{subsec:infrastructure}

All experiments were conducted on high-performance computing platforms optimized for large-scale deep learning tasks. To support different model architectures and robustness checks, we utilized a hybrid hardware setup tailored to specific backbones:

\begin{itemize}
    \item \textbf{Hardware Environment:}
    \begin{itemize}
        \item \textbf{Huawei Ascend 910B NPU:} Dedicated to serving the \texttt{pangu\_embedded\_1b} model.
        \item \textbf{Nvidia RTX A6000 GPU:} Dedicated to serving the \texttt{Meta-Llama-3.1-8B-Instruct} model.
        \item \textbf{Storage:} 50 GB+ available disk space for dataset management and model weight storage.
    \end{itemize}
    
    \item \textbf{Software Environment:}
    \begin{itemize}
        \item \textbf{Operating System:} Linux.
        \item \textbf{Programming Language:} Python 3.11 or higher.
        \item \textbf{Core Models:}
        \begin{itemize}
            \item[1.] \textbf{Embedding Models:}
            \begin{itemize}
                \item \texttt{BAAI/bge-m3}: Utilized as the default model for high-dimensional semantic representation.
                \item \texttt{intfloat/e5-large-v2}: Utilized for robustness verification experiments.
            \end{itemize}
            \item[2.] \textbf{Language Models (LLM):}
            \begin{itemize}
                \item \texttt{pangu\_embedded\_1b}: The primary backbone generative model, executed on the Ascend 910B platform.
                \item \texttt{Meta-Llama-3.1-8B-Instruct}: Employed for cross-model validation and robustness checks, executed on the Nvidia A6000 platform.
            \end{itemize}
        \end{itemize}
    \end{itemize}
\end{itemize}

\subsection{Implementation and Workflow Details}
\label{subsec:workflow}

We are currently organizing the source code and experimental scripts to ensure long-horizon usability. Upon publication, we plan to release the full implementation as an open-source repository and a Python package via PyPI. This package is designed to facilitate direct streaming evaluation, allowing third-party researchers to decompose their custom memory modules and deploy them onto our evaluation pipeline for granular assessment. The standardized workflow is structured as follows:

\begin{enumerate}
    \item \textbf{Environment Setup:} The framework requires a Conda environment with Python 3.11. Future deployment will be streamlined via standard package managers:
    \newline
    \texttt{conda create -n mem python=3.11}
    \newline
    \texttt{pip install neuromem-eval (anonymous link provided in supplementary)} 

    \item \textbf{Dataset Acquisition:} We evaluate on the \texttt{locomo} dataset. The framework includes utilities to automatically download and format these benchmarks for streaming protocols:
    \newline
    \texttt{neuromem-data download locomo}

    \item \textbf{Configuration:} System behaviors are fully configurable. Users can define model specifications and pipeline parameters (e.g., \texttt{benchmarks/experiment/config}) to customize the lifecycle dimensions (D1-D5).

    \item \textbf{Execution \& Analysis:} The core logic orchestrates the interleaved insertion and retrieval process. The suite separates execution (\texttt{benchmarks/experiment/script}) from analysis (\texttt{benchmarks/evaluation/scripts}), ensuring that metrics are computed consistently across different system instantiations.
\end{enumerate}

\subcompact
\section{Detailed Experimental Results}
\label{app:detailed_experimental_results}

\subsection{Extended Analysis}
\label{app:extended_analysis}

\subsubsection{Scalability Analysis of Iterative Graph Memory}
\label{app:hipporag_scalability}

While graph-based architectures like HippoRAG demonstrate superior multi-hop reasoning capabilities, our analysis reveals significant scalability bottlenecks when applied to streaming protocols. Streaming deployment requires continuous graph maintenance, exposing the computational cost of iterative topology updates. Our profiling decomposes the insertion latency into two primary components: \textit{Information Extraction Overhead} and \textit{Synonymy Edge Construction}.

\textbf{Information Extraction Overhead.} 
Before graph construction, each incoming memory segment undergoes OpenIE processing, involving Named Entity Recognition (NER) and Triple Extraction. Since a single memory segment typically spawns multiple entities and facts, this step incurs a substantial fixed latency. Even with optimized local LLMs, the sequential execution of these generative calls imposes a base overhead of approximately 5-10 seconds per insertion, regardless of the memory size. This high entry cost makes sub-second real-time interaction challenging even at the start of the lifecycle.

\begin{figure}[h]
    \centering
    \includegraphics[width=0.6\textwidth]{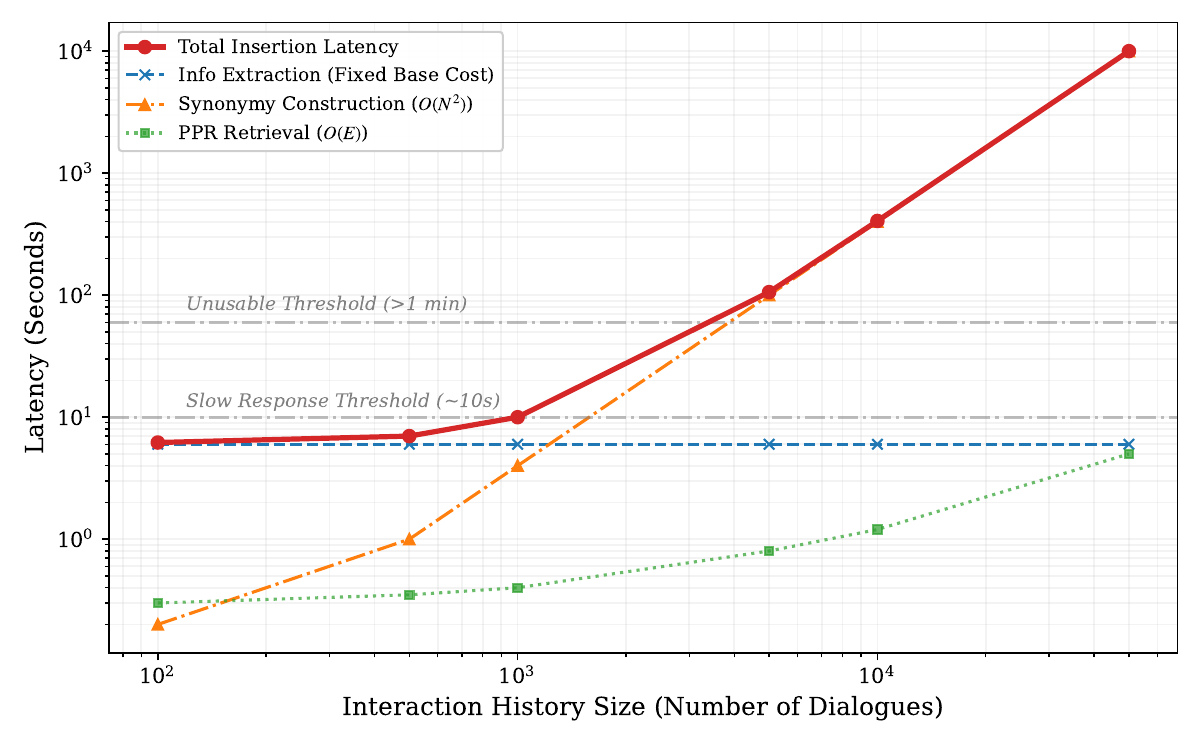}
    \caption{\textbf{Scalability Analysis of Iterative Graph Memory.} The log-log plot decomposes insertion latency into fixed and dynamic components. The \textit{Information Extraction} imposes a high initial latency floor due to LLM calls. As memory scales, the \textit{Synonymy Edge Construction} dominates with quadratic growth, causing total latency to exceed 1 minute beyond 5,000 dialogues. The \textit{PPR Traversal} remains relatively efficient but adds to the retrieval latency.}
    \label{fig:hipporag_scalability}
\end{figure}

\textbf{Synonymy Edge Construction (Quadratic Growth $O(N^2)$).} 
As the graph grows, the maintenance of semantic consistency becomes the dominant bottleneck. Identifying synonymy edges relies on a global K-Nearest Neighbors (KNN) search. Critically, the number of entities $N$ grows significantly faster than the number of dialogue turns due to the multiplicative factor of extraction. At each insertion step $t$, the system computes pairwise similarity scores for $N_t$ accumulated entities, resulting in $O(N_t^2 \cdot d)$ complexity. 
As illustrated in Figure~\ref{fig:hipporag_scalability}, while the Information Extraction cost remains constant, the Synonymy Construction cost exhibits quadratic growth. For a history of 5,000 dialogues, the insertion latency explodes to over 100 seconds, rendering the system unusable for online streaming.

\textbf{Contextual Integration ($O(I \cdot |E|)$).} 
Following graph updates, retrieval executes the Personalized PageRank (PPR) algorithm. While efficient for sparse graphs, the iterative convergence still imposes a latency floor of approximately 0.5 to 1.0 seconds for mid-sized graphs ($N \approx 25,000$), further straining the total response time.

In conclusion, iterative graph memory faces a two-fold scalability barrier: a high constant latency floor due to generative extraction and an exponential quadratic latency surge due to non-incremental graph maintenance. Effective streaming deployment requires asynchronous extraction pipelines and strictly incremental graph update algorithms (e.g., local neighborhood search) to decouple maintenance costs from total memory size.

\subsection{Robustness Experiment}

\subsubsection{Robustness of Normalization Strategy}
\label{app:robustness_embedding}

To rule out the possibility that the poor performance of the \texttt{Rewrite} strategy was due to limited model capacity, we replicated the experiment on LoCoMo using the stronger \texttt{intfloat/e5-large-v2} embedding model. 

\begin{figure}[h]
    \centering
    \includegraphics[width=0.6\textwidth]{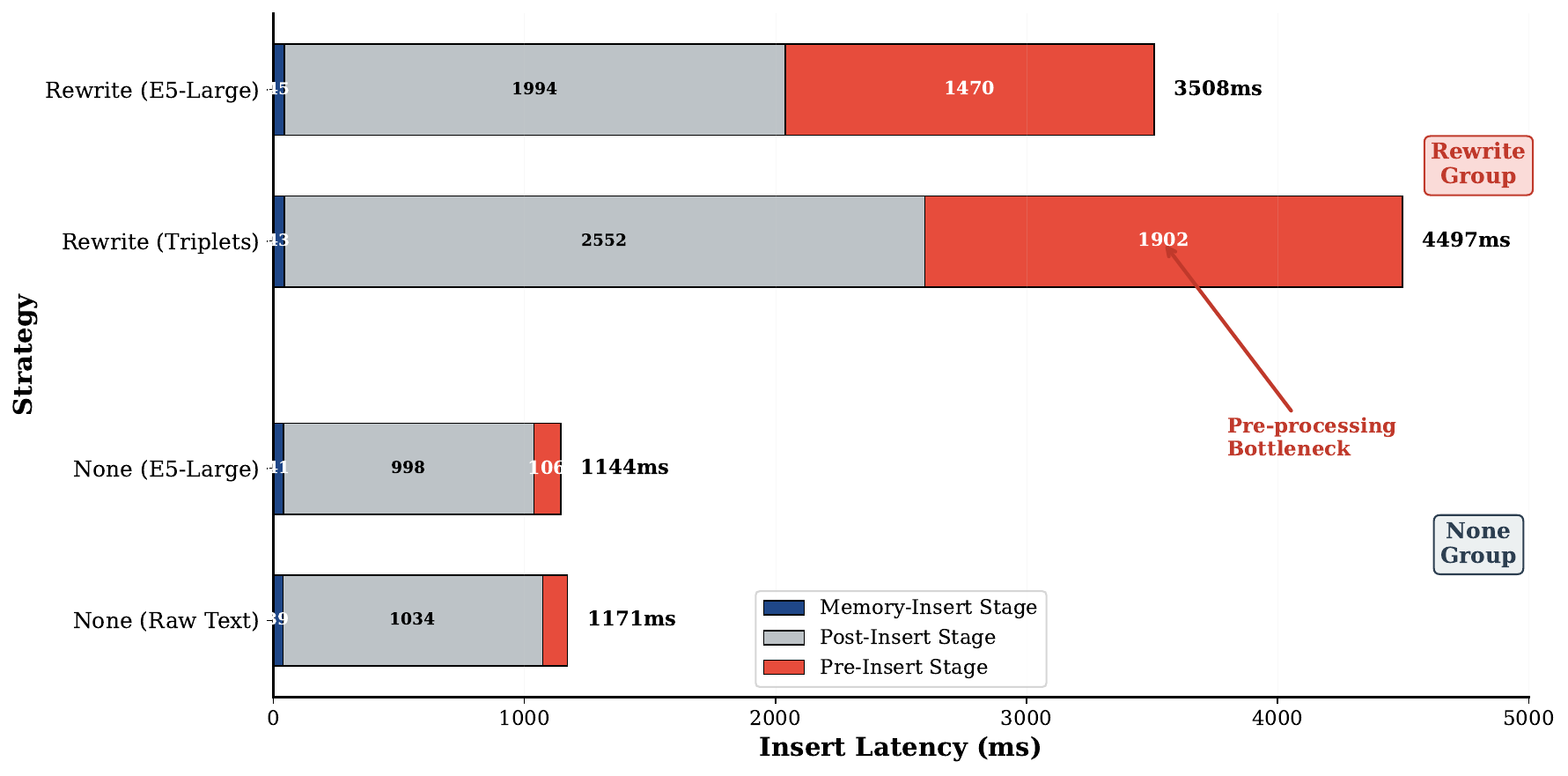}
    \caption{\textbf{Insert Latency Breakdown by Normalization Strategy.} The horizontal bars compare the time cost of ingestion stages. The \texttt{Rewrite} strategy incurs a massive bottleneck, taking nearly 2 seconds just to extract triplets before any storage operation occurs. In contrast, the \texttt{None} baselines have negligible pre-processing costs.}
    \label{fig:normalization_latency}
\end{figure}

\begin{table}[h]
    \centering
    \caption{\textbf{Robustness Check on Normalization with E5-Large.} Even with a superior embedding backbone, the \texttt{Rewrite} strategy consistently underperforms the \texttt{None} baseline across all rounds. The latency breakdown reveals a heavy preprocessing cost for triplet extraction.}
    \label{tab:e5_robustness_wide}
    
    \small 
    
    \begin{tabular*}{\textwidth}{@{\extracolsep{\fill}} l ccccc cc}
    \toprule
    \multirow{2}{*}{\textbf{Normalization}} & \multicolumn{5}{c}{\textbf{Token-level F1 per Round}} & \multirow{2}{*}{\textbf{Mean F1}} & \multirow{2}{*}{\textbf{Insert Latency}} \\
    \cmidrule(lr){2-6}
    & \textbf{R1} & \textbf{R2} & \textbf{R3} & \textbf{R4} & \textbf{R5} & & \\
    \midrule
    \textbf{None (Raw Text)} & \textbf{0.326} & \textbf{0.332} & \textbf{0.320} & \textbf{0.319} & \textbf{0.297} & \textbf{0.319} & \textbf{1,171\,ms} \\
    Rewrite (Triplets) & 0.168 & 0.128 & 0.120 & 0.116 & 0.111 & 0.129 & 4,497\,ms \\
    \midrule
    \textit{$\Delta$ vs. Baseline} & -48.5\% & -61.4\% & -62.5\% & -63.6\% & -62.6\% & -59.7\% & +284\% (Slower) \\
    \bottomrule
    \end{tabular*}
\end{table}

The results, presented in Table~\ref{tab:e5_robustness_wide} and visualized in Figure~\ref{fig:normalization_latency}, confirm that the performance gap is fundamental to the structural representation rather than the embedding quality. The \texttt{Rewrite} strategy suffers a severe drop in F1 score (Mean $\Delta \approx -60\%$) compared to the raw text baseline. 

Crucially, the latency breakdown in Figure~\ref{fig:normalization_latency} exposes the root cause of the inefficiency. While the core memory insertion time is negligible for all methods, the \texttt{Rewrite} strategy is dominated by the \textbf{Normalization Stage}, which demands approximately 1,902 ms to extract triplets from the raw text. This structural overhead results in a total insertion latency that is nearly $4\times$ slower than the baseline (4,497 ms vs 1,171 ms). This reinforces our conclusion that for embedding-based retrieval, preserving the rich semantic texture of raw text is strictly superior to aggressive structural normalization, both in terms of accuracy and efficiency.

\subsubsection{Robustness of Context Integration Strategy}
\label{app:robustness_integration}

To investigate whether the high latency of generative fusion was merely an artifact of our default backbone's inference speed (pangu-1b), we replicated the Context Integration experiment using the faster and more capable \textbf{Llama-3-8B} model.

\begin{figure}[h]
    \centering
    \includegraphics[width=\textwidth]{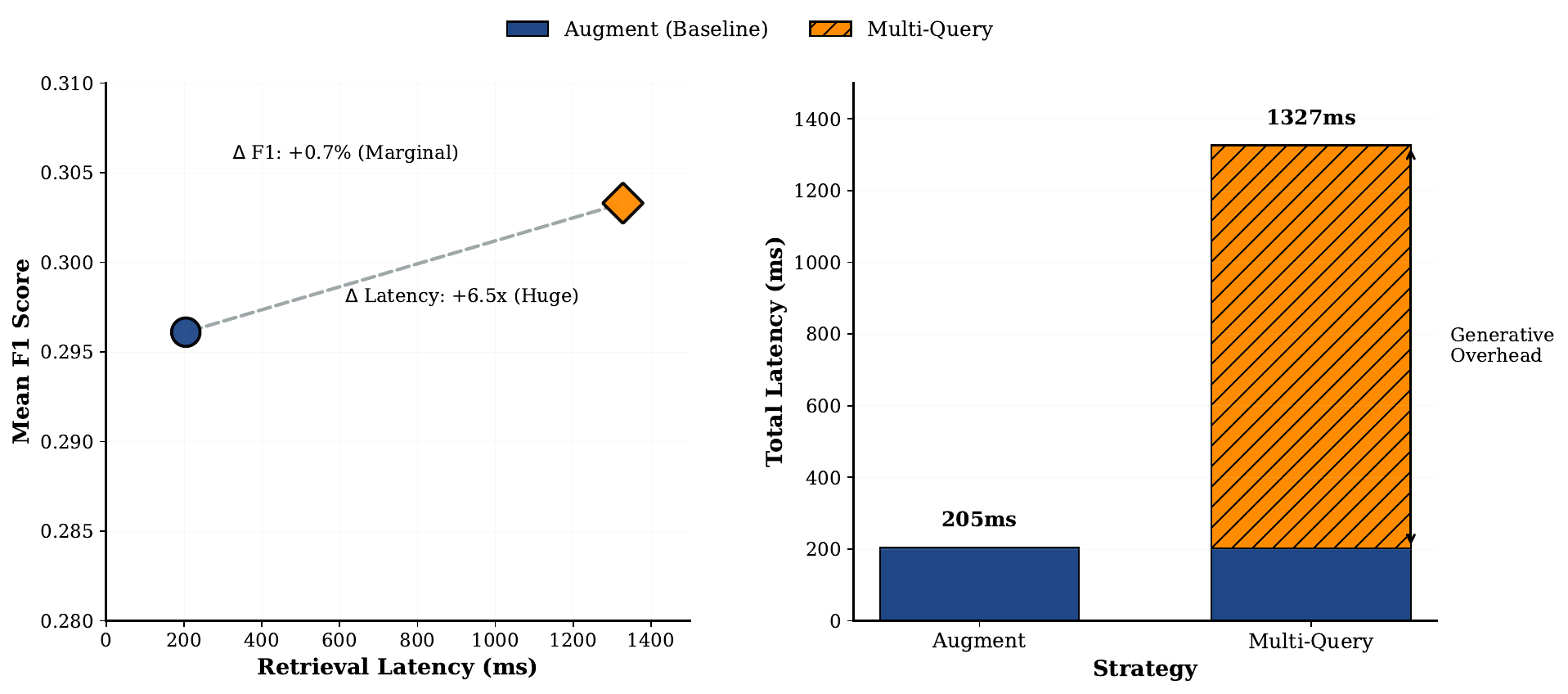}
    \caption{\textbf{Efficiency-Accuracy Trade-off with Llama-3-8B.} The left panel illustrates the disproportionate cost of generative fusion: a marginal mean F1 gain (+0.7\%) requires a massive $6.5\times$ increase in latency. The right panel decomposes this latency, revealing that the cost is driven almost entirely by the Generative Overhead (1.1s) during the Context Integration stage, while the base retrieval time remains constant.}
    \label{fig:llama_integration_tradeoff}
\end{figure}

\begin{table}[h]
    \centering
    \caption{\textbf{Robustness Check on Context Integration with Llama-3-8B.} Performance across all 5 rounds confirms that \texttt{Multi-Query} yields consistent but marginal gains. The latency breakdown highlights that the cost is driven entirely by the generative Context Integration stage, which adds over 1 second of overhead compared to the \texttt{Augment} baseline.}
    \label{tab:llama_integration_robustness}
    
    \small
    
    \begin{tabular*}{\textwidth}{@{\extracolsep{\fill}} l ccccc c cc}
    \toprule
    \multirow{2}{*}{\textbf{Integration Strategy}} & \multicolumn{5}{c}{\textbf{Token-level F1 per Round}} & \multirow{2}{*}{\textbf{Mean F1}} & \multicolumn{2}{c}{\textbf{Retrieval Latency Breakdown}} \\
    \cmidrule(lr){2-6} \cmidrule(lr){8-9}
    & \textbf{R1} & \textbf{R2} & \textbf{R3} & \textbf{R4} & \textbf{R5} & & \textbf{Context Integration} & \textbf{Total} \\
    \midrule
    \texttt{Augment} (Baseline) & 0.339 & 0.301 & 0.289 & 0.283 & 0.268 & 0.296 & $\sim$0.3\,ms & \textbf{205\,ms} \\
    \texttt{Multi-Query} & \textbf{0.348} & \textbf{0.306} & \textbf{0.295} & \textbf{0.289} & \textbf{0.279} & \textbf{0.303} & 1,125\,ms & 1,328\,ms \\
    \midrule
    \textit{Impact ($\Delta$ vs Baseline)} & \textbf{+2.6\%} & \textbf{+1.7\%} & \textbf{+2.1\%} & \textbf{+2.1\%} & \textbf{+4.1\%} & \textbf{+2.4\%} & \textbf{Massive Overhead} & \textbf{6.5$\times$ Slower} \\
    \bottomrule
    \end{tabular*}
\end{table}

The results, visualized in Figure~\ref{fig:llama_integration_tradeoff} and detailed in Table~\ref{tab:llama_integration_robustness}, indicate that the efficiency bottleneck is structural rather than model-dependent. Expanding the evaluation to the full five-round lifecycle reveals that while the \texttt{multi\_query} strategy consistently outperforms the baseline—raising the Mean F1 score from 0.296 to 0.303—this advantage is slight. The strategy achieves a 4.1\% relative improvement in the final round, but this accuracy benefit comes at a disproportionate cost. The total retrieval latency spikes from a manageable 205 milliseconds in the heuristic \texttt{augment} baseline to 1,328 milliseconds with generative fusion, representing a 6.5-fold increase in processing time.

Crucially, the granular breakdown identifies the root cause of this bottleneck. As shown in the right panel of Figure~\ref{fig:llama_integration_tradeoff}, while the baseline requires less than 1 millisecond for context integration, the generative approach demands approximately 1,125 milliseconds to synthesize and rerank the retrieved contexts. Consequently, the Context Integration stage alone accounts for the vast majority of the total latency. This finding reinforces our conclusion that generative integration acts as a ``latency tax'': upgrading the model backbone reduces absolute inference time but fails to eliminate the massive architectural overhead required for generative fusion.

\end{document}